\documentclass[letterpaper]{article}
\usepackage{gsignatures}
\usepackage{times}
\usepackage{helvet}
\usepackage{courier}
\usepackage[hyphens]{url}
\usepackage{graphicx}
\usepackage{natbib}
\usepackage{caption}
\usepackage{algorithm}
\usepackage{algorithmic}

\usepackage{newfloat}
\usepackage{listings}
\DeclareCaptionStyle{ruled}{labelfont=normalfont,labelsep=colon,strut=off}
\floatstyle{ruled}
\newfloat{listing}{tb}{lst}{}
\floatname{listing}{Listing}

\usepackage{ml_defs}

\usepackage{nicefrac}

\usepackage[capitalize,noabbrev]{cleveref}

\usepackage{algorithm}
\usepackage{algorithmic}

\usepackage{caption}
\usepackage{subcaption}

\usepackage{booktabs}
\usepackage{siunitx}
\usepackage{makecell}
\usepackage{multirow}
\usepackage{selectp}

\title{G-Signatures: Global Graph Propagation With Randomized Signatures}
\author{
    Bernhard Sch\"{a}fl\equalcontrib \textsuperscript{\rm 1} \quad
    Lukas Gruber\equalcontrib \textsuperscript{\rm 1} \quad
    Johannes Brandstetter\textsuperscript{\rm 3} \quad
    Sepp Hochreiter\textsuperscript{\rm 1,\rm 2}
}
\affiliations{
    \textsuperscript{\rm 1}ELLIS Unit Linz and LIT AI Lab, Institute for Machine Learning, Johannes Kepler University Linz, Austria\\
    \textsuperscript{\rm 2}Institute of Advanced Research in Artificial Intelligence (IARAI)\\
    \textsuperscript{\rm 3}Microsoft Research AI4Science
 }
\begin{document}

\maketitle

\begin{abstract}
        Graph neural networks (GNNs) have evolved into one of the most popular
        deep learning architectures.
        However, GNNs suffer from over-smoothing node information and, therefore,
        struggle to solve tasks where global graph properties are relevant. 
        We introduce G-Signatures, a novel graph learning 
        method that enables global graph propagation via randomized signatures.
        G-Signatures use a new graph conversion concept
        to embed graph structured information 
        which can be interpreted as paths in latent space.
        We further introduce the idea of latent space path mapping. This
        allows us to iteratively traverse latent space paths, 
        and, thus globally process information.
        G-Signatures excel at extracting and processing global graph properties, 
        and 
        effectively
        scale to large graph problems.
        Empirically, we confirm the advantages of 
        G-Signatures 
        at 
        several
        classification and regression tasks.
\end{abstract}

    \section{Introduction}

        Graph neural networks (GNNs), like 
        graph convolutional networks~\citep{Kipf:17}, 
        GraphSAGEs~\citep{Hamilton:17},
        graph attention networks~\citep{Velickovic:18},
        or message passing GNNs~\citep{Gilmer:17}
        are one of the most popular and most successful deep learning architectures.
        GNNs are based on the principle of learning interactions between many entities
        in forward dynamics~\citep{Battaglia:18}
        and, therefore, advance physical simulations of molecular modeling, 
        fluid dynamics, weather forecasting, and aerodynamics
        \citep{batatia:22, Li:18particle, SanchezGonzalez:20, Pfaff:20, Mayr:21, Brandstetter:22a, Keisler:22, Lam:22}.
        However, the multi-layer message aggregation scheme of GNNs is prone
        to over-smoothing node information
        even after a few propagation steps~\citep{Li:18,Chen:20,Zhu:21,Alon:21},
        yielding node representations that tend to be similar to each other.
        Furthermore, since GNNs usually process information 
        as messages across edges,
        the number of messages grows exponentially with the width of a GNN's receptive field, 
        resulting in 
        over-squashing
        for more propagation steps~\citep{Zhu:20,Chen:20,Alon:21}.
        Consequently, whole-graph classification and regression tasks that 
        comprise long-range dependencies are very
        challenging for GNNs~\citep{Xu:18}.
        There has been extensive work to mitigate effects of over-smoothing and
        over-squashing~\citep{Chen:19,Hu:20,Liu:20_1,Gu:20,Yang:21,Kim:21,Liu:22,Alon:21,Rampasek:22}.
        Prominent examples of such are Gated GCNs \citep{Bresson:17}
        or Graph Transformers \citep{Dwivedi:21},
        which are amongst the best performing
        methods for longer-range graph interactions \citep{Dwivedi:22,Rusch:23,Rusch:23_1,Rusch:22}.
        However, all these GNN variants are still underperforming when predicting global 
        properties of graphs.
        
        \begin{figure*}[ht]
            \begin{center}
            \includegraphics[width=\textwidth]{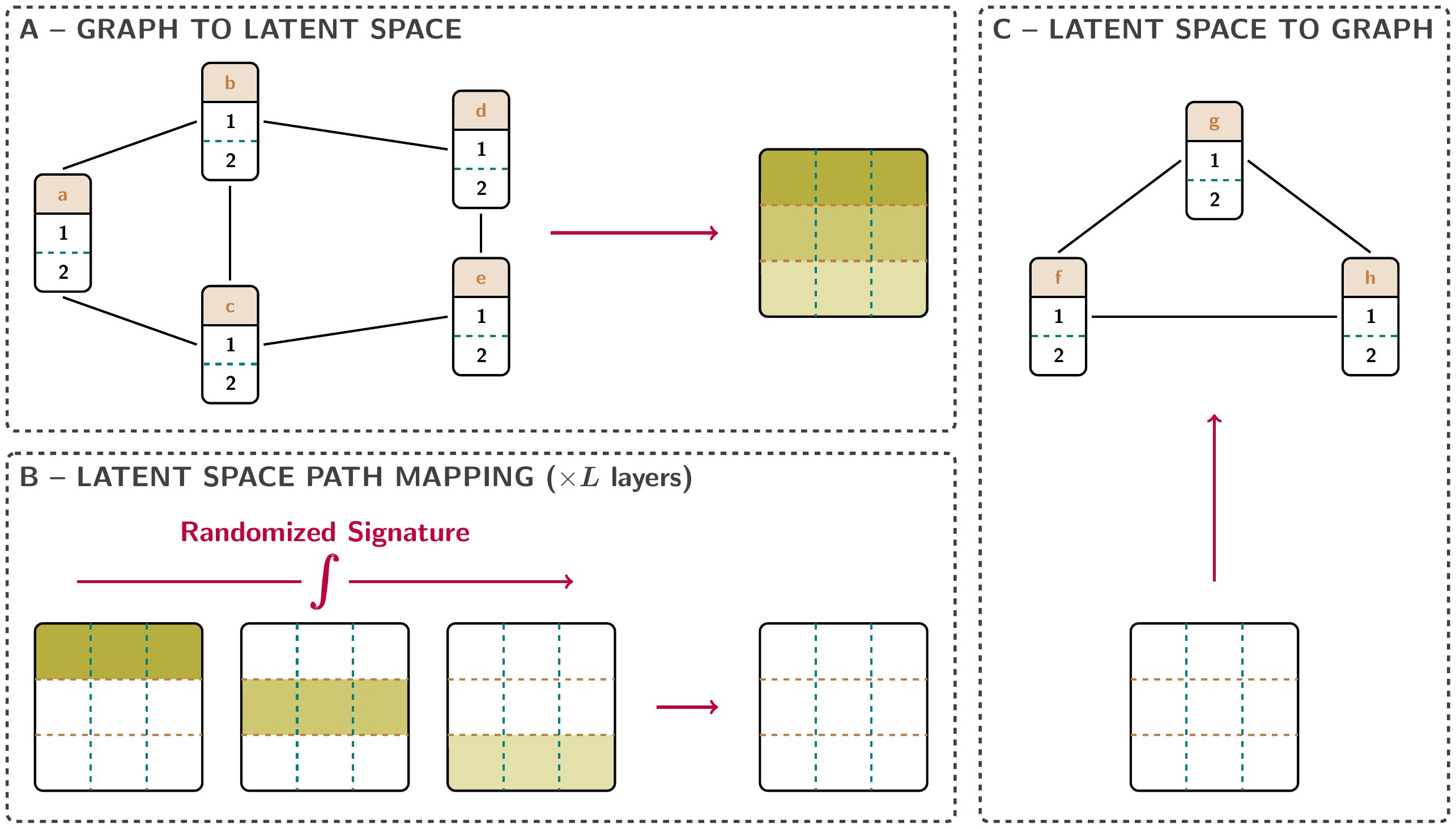}
            \caption{
            Schematic sketch of G-Signatures.
            G-Signatures apply randomized signature layers in latent space, which enables 
            to globally and efficiently process graphs.
            \textit{(A)} Graph structured data that is converted and 
            interpreted as ``path'' along one dimension in latent space.
             \textit{(B)} Latent space path mappings (LSPM) via randomized signature layers 
             which traverse the converted graph iteratively on a per-feature basis. 
             \textit{(C)} The latent path is mapped to the target representation.}\label{fig:rsnet_overview}
             \end{center}
        \end{figure*}
        
        In this work, we introduce G-Signatures,
        which efficiently learn randomized signatures
        to solve various graph tasks via gradient descent.
        Most notably, G-Signatures enable global graph propagation since at their core 
        G-Signatures replace
        the local concept of message aggregation by globally traversing the graph.
        We achieve this by interpreting
        graphs
        as paths and calculate the signature thereof.
        Roughly speaking, a path is a mapping from an interval into a space where
        its \emph{signature} describes the path uniquely
        in an efficient, computable way~\citep{Chen:54,Chen:57,Chen:58}.
        However, the signature cannot be computed directly since it is an infinite sequence.
        \emph{Truncated signatures} approximate this sequence by cutting at a specific level. G-Signatures use \emph{randomized signatures},
        which approximate truncated signatures via random feature mappings.
        Truncated signatures
        have been suggested as a layer 
        in neural networks to represent input data ~\citep{Kidger:19}.
        Furthermore, randomized signatures have already been used for 
        differential equations~\citep{Cuchiero:21}, and for
        time series prediction~\citep{Compagnoni:22}.
        Signatures have been used for (character) recognition tasks in
        machine learning \citep{Yang:15,Li:17_1,Xie:18,Yang:17}, 
        as well as for data streams and pricing in finance \citep{Lyons:14,Lyons:19,Kalsi:19}, 
        and for rough paths \citep{Friz:10,Lyons:98}.
        Likewise,~\cite{Toth:22} model the local neighborhood of a given node in a GNN by interpreting it as a path over nodes with a limited path length, and summarize the path information with truncated signatures.
        In contrast to these approaches, 
        we use randomized signatures
        to represent information 
        as a path
        over the features 
        of the nodes in a graph. 
        G-Signatures learn how to refine the random mappings via gradient descent, that is,
        they learn to extract task-relevant information from the signature representation of a path.
        In doing so, we are, to the best of our knowledge, the first 
        to present a framework for learning to extract signature information
        from 
        graph structured data.        
        In this work we introduce two novel concepts:
        (i) \emph{graph conversion}, i.e., 
        an efficient merging and embedding of graph information which we interpret as a path in latent space, and 
        (ii) \emph{latent space path mappings} (LSPM), i.e., 
        new layers that map from one path to another in latent space.
        These two concepts constitute \emph{G-Signatures},
        which is a novel deep learning architecture with its own learning algorithm
        sketched in \cref{fig:rsnet_overview}. 
        G-Signatures resemble GNNs, but in contrast to GNNs
        they offer an efficient and scalable solution 
        to whole-graph classification and regression tasks.
        G-Signatures are a paradigm shift from collecting information locally 
        for each node towards collecting 
        information along paths through the graph.
        In G-Signatures, 
        the signature transform is used in a similar way as 
        Fourier analysis on Euclidean domains~\citep{Li:20fno} and, most notably,
        as spectral graph theory which can be seen as 
        Fourier analysis on non-Euclidean domains~\citep{Bronstein:17, Bronstein:21}.
        The largest disadvantage of spectral graph theory
        is that the eigendecomposition of the graph Laplacians is computationally expensive,
        especially for large graphs~\citep{Defferrard:16, Kipf:17}. 
        In contrast, G-Signatures allow for efficient 
        layer-wise propagation of global graph information. 
        To summarize, the contributions of this paper are: 
        \begin{itemize}
        \item We introduce G-Signatures for global graph propagation based on randomized signatures. 
        \item We establish the concepts of graph conversion and latent space path mappings (LSPM) 
        which enable us to apply randomized signatures to graph structured data.
        \item In experiments, we demonstrate that G-Signatures
        (i) are able to learn global graph characteristics, 
        (ii) offer an efficient and scalable solution to large graph problems,
        and (iii) are a more generally applicable method not limited to typical GNN tasks.
        \end{itemize}
                        
        \section{Background and Related Work}
        In this section we discuss graph structured data, the randomized signature transform, and 
        graph conversion.
        
        \textbf{Learning on graph structured data.}
        We consider undirected graphs $\cG = (\cV,\cE)$ 
        with nodes $v_i \in \cV$, and edges $e_{ij} \in \cE$, 
        where $d$-dimensional node features $\Bh_{i} \in \dR^{d}$
        are attached to each of the nodes.
        Whether an edge between a pair of nodes $(v_i, v_j)$
        is contained in the graph $\cG$ depends 
        on the connectivity criterion between two nodes. 
        For example, if distance is chosen as criterion, we might insert an edge when 
        the cut-off radius $r_{\text{cut-off}}$ is below a threshold:
        $
            e_{ij} \in \cE \iff d(v_i, v_j) \leq r_{\text{cut-off}} \ .
        $
        The connectivity is summarized in the adjacency matrix $\bm{A} \in \dR^{N \times N}$, 
        which can be binary or weighted.
        Graph neural networks (GNNs)~\citep{Scarselli:08, Battaglia:18}
        are designed to learn from graph structured data and are by construction permutation 
        equivariant with respect to the input. 
        
        \textbf{Path.}
        A \emph{path} $X : [a, b] \rightarrow \dR^d$ is a continuous mapping 
        from an interval $[a, b]$ to $\dR^d$~\citep{Chevyrev:16}. 
        For a given path, the \emph{signature} of the path summarizes 
        its statistics (see~\cref{app:moments}), 
        whereas the mapping from a path to its signature is the \emph{signature transform}.
        
        \textbf{Signature (transform) of a path. } 
        For a path $X: [a,b] \rightarrow \dR^d$ 
        with coordinate paths $X^1_t,\ldots,X^d_t$,
        where each $X^i [a,b] \rightarrow \dR$ is a real-valued path, the signature transform 
        as originally introduced by~\citet{Chen:54,Chen:57,Chen:58} is defined as:
        \begin{align}\label{eq:level_k}
                S(X)^{i_1, \ldots , i_{k}}_{a,t} \ &:= \ \int\displaylimits_{a < t_{k} < t}{\ldots} \int\displaylimits_{a < t_1 < t_{2}}{\Rd X^{i_{1}}_{t_{1}} \ldots \Rd X^{i_{k}}_{t_{k}}} \ ,
        \end{align}
        where multi-index $i_1, \ldots , i_{k} \in \{1, \ldots, d \}^k$ and
        \begin{align}
                S(X)^{i_1}_{a,t} \ &:= \ \int\displaylimits_{a < s < t} \Rd X^{i_1}_{s} 
                \ = \ X^{i_1}_{t} \ - \ X^{i_1}_{0}
        \end{align}
        integrates each 
        dimension
        separately, i.e., computes the per-coordinate
        increments. 
        The pointwise evaluation $S(X)^{i_1}_{a,\cdot}: [a,b] \rightarrow \dR$ is again a path. 
        The collection of all feature increments concludes the first level of the signature. 
        The second level is defined as:
        \begin{align}
        \label{eq:double_iterated_integral}
                S(X)^{i_1,i_2}_{a,t} \ :&= \ \int\displaylimits_{a < s < t} S(X)^{i_1}_{a,s} \ \Rd X^{i_2}_{s} \\ &= \ \int\displaylimits_{a < r < s < t} \Rd X^{i_1}_{r} \ \Rd X^{i_2}_{s} \ ,
        \end{align} 
        with all further levels defined likewise. The double iterated integral of \cref{eq:double_iterated_integral} integrates one path, i.e., $S(X)^{i_1}$ against the other $X^{i_2}$.
        The collection of all iterated integrals of a path $X$ in the interval $[a, b]$ is called the \emph{signature} of the path $X$, and is denoted as $S(X)_{a, b}$~\citep{Cuchiero:21}:%
        \begin{equation}
            \begin{split}
                S(X)_{a,b} \ := \ (1, &S(X)^{1}_{a,b}, \dots , S(X)^{d}_{a,b},\\&S(X)^{1, 1}_{a,b}, \ldots , S(X)^{d, d}_{a,b}, \ldots)\ .
            \end{split}
            \label{eq:signature}
        \end{equation}
        
        The signature itself exhibits a universal non-linearity property, 
        that is linear functionals on the signature are dense in the set of functions on $X$ \citep{Arribas:18}.
        An informal excerpt of this result is the following: A continuous function 
        $f(X)$ of a path $X$ can be approximated with an error lower or equal to  $\epsilon$
        by a linear mapping $L$ of the signature $S(X)$:
        \begin{align}
            \forall \epsilon > 0 \textbf{ } \exists L : \Vert f(X) - L(S(X)) \Vert \leq \epsilon\ .
        \end{align}
        Hence, the signature of a path $X \in \mathbb{R}^{d}$ acts as a reservoir in terms of reservoir computing~\citep{Cuchiero:21},
        that is the signature acts as a fixed non-linear system and maps input signals into higher dimensional computational spaces.
        As a result, the signature gives a basis representation of functions on $X$ in the Euclidean domain.

        \textbf{Truncated signature. } 
        The signature of \cref{eq:signature} itself cannot be computed, as it is an infinite sequence. Cutting this sequence at a specific level yields the truncated signature at level $M$:
        \begin{align}
            S(X)^{M}_{a,b} \ &:= \ \left(1,S(X)^1_{a,b}, 
            \dots ,S(X)^{d,\dots ,d}_{a,b} \right) \ ,
        \end{align}
        where the superscript of the last entry is the multi-index denoting $M$-times the $d$-th coordinate 
        of the path $X\in \dR^{d}$ analog to~\cref{eq:level_k}.
        In general, the number of signature terms at level $M$ is
        \begin{align}
            \left|S(X)^{M}_{a,b}\right| \ &= \ (d^{M + 1} - 1)/(d - 1) \ .
        \end{align}
        
        \textbf{Randomized signature. }
        The randomized signature is based on the Johnson-Lindenstrauss Lemma \citep{Johnson:84},
        which describes a distance-preserving, low-dimensional embedding of high-dimensional data.
        For a path $X:[0,T] \rightarrow \dR^d$         
        where $X$ consists of $d$ coordinate paths sampled at $N$ points $1 < \ldots < N=T$~\citep{Compagnoni:22}.
        The $k$-dimensional randomized signature
        $S_R(X)$ of $X$ is the vector $\bm{z}_T \in \dR^{k}$
        which is obtained via $N$ incremental updates.
        $\bm{z}_T$ approximates the signature of path $X$, with a vanishing error as $k \rightarrow \infty$. For more details see Theorem~3.3 of \citet{Compagnoni:22}. 
        In the following, we refer to $\bm{z}_T$ and the respective $N-1$ preceding incremental update steps, i.e., $\left(\bm{z}_1,\ldots, \bm{z}_T\right)^T$ as \emph{randomized signature matrix} $\bm{Z} \in \dR^{N \times k}$, which again is a path itself.
        A transposed and for G-Signatures modified computation can be seen in \cref{alg:randomized} with a path
        $X:[0,T] \rightarrow \dR^N$ sampled at $d$ points $1 < \ldots < d=T$.

        \begin{algorithm}[ht]
           \caption{Randomized Signature Layer}\label{alg:randomized}
            \begin{algorithmic}[1]
               \REQUIRE Matrix $\bm{X} \in \mathbb{R}^{d \times N}$ represents a path sampled at $d$ points $1 < \ldots{} < d=T$, 
               randomized signature size $k$, activation function~$\sigma$, and number of signature heads $p$
               \vspace{1pt}
               \ENSURE $\bm{z}_{0} \in \mathbb{R}^{k}$, $\bm{A}_{i} \in \mathbb{R}^{k \cdot p \times k}$, $\bm{b}_{i} \in \mathbb{R}^{k \cdot p}$ from $\mathcal{N}(0, \nicefrac{1}{k})$, \\ and $\bm{W} \in \mathbb{R}^{k\times{}k\cdot{}p}$, $\bm{o} \in \mathbb{R}^{k}$ from $\mathcal{U}(\nicefrac{-1}{\sqrt{k\cdot{}p}}, \nicefrac{1}{\sqrt{k\cdot{}p}})$.
               \vspace{1pt}
               \FOR{$j = 1$ {\bfseries to} $d$}
                   \STATE $\delta{}\bm{z}_j \gets \bm{W} \big( \sum_{i = 1}^{N}\ \sigma{\big(\bm{A}_{i}\bm{z}_{j-1}^{} + \bm{b}_{i}\big)}X_j^i \big) + \bm{o}$
                   \vspace{1.2pt}
                   \STATE $\bm{z}_{j} \gets \bm{z}_{j-1} +
                   \delta{}\bm{z}_j$
                   \vspace{1.2pt}
               \ENDFOR
            \end{algorithmic}
        \end{algorithm}
        The computation of the randomized signature can
        be 
        seen
        as 
        being 
        part of the broad family of gated deep neural architectures, such as LSTMs~\citep{Hochreiter:91a,Hochreiter:96a,Hochreiter:97}, GRUs~\citep{Cho:14}, or Highway Networks \citep{Srivastava:15}.
        
\begin{figure*}[t]
            \begin{center}
            \includegraphics[width=\textwidth]{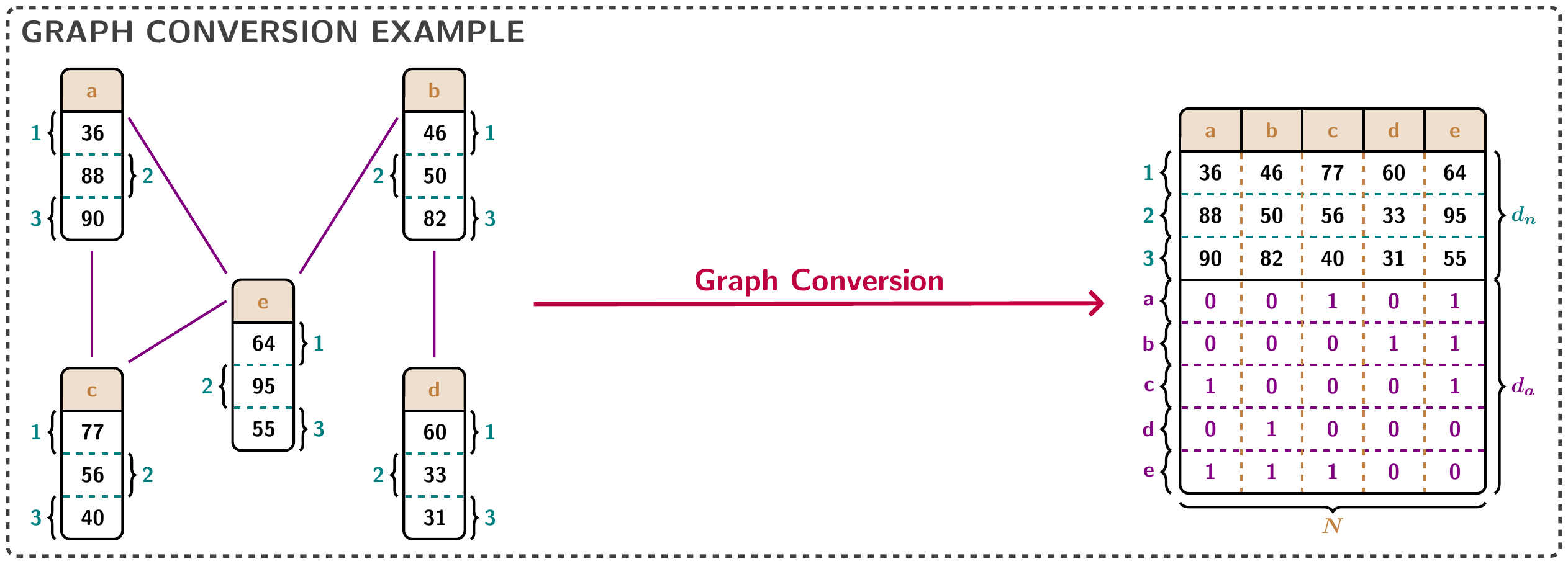}
            \caption{Example showcasing our proposed graph conversion concept. Graph conversion allows us to efficiently compress graph structured data into a latent representation while preserving node as well as adjacency information. For visual clarity the latent adjacency information is depicted in a simplified way where it is actually computed according to \cref{eq:embeddings}.}\label{fig:graph_as_path}
            \end{center}
        \end{figure*}

        \textbf{Randomized signature of graphs. }
        For a given graph $\cG = (\cV,\cE)$ with $N$ nodes and node features $\bm{h}_i \in \dR^{d}$
        attached to them, we can interpret the graph as a path $G:[0,T] \rightarrow \dR^N$ where $G$ 
        consists of $N$ coordinate paths sampled
        at $d$ points $1 < \ldots < d=T$. 
        Put differently, time step $j$ in the path consists of features
        $\{\bm{h}_{i,j}\}_{i=1}^N$ taken from all $N$ nodes in the graph.
        The path is defined over features instead of nodes in order to keep favorable GNN properties like permutation invariance of processing node information. This also enables the processing of global information of the entire graph in each step without depending on the concept of local information aggregation as in Message Passing GNNs (MPNNs).
        In \cref{sec:globprop} we give further intuition based on two important examples where the ability to propagate global graph information is vital to solve the given tasks.
        When interpreting a path as a random variable \cite{Chevyrev:16,Chevyrev:18} have shown that the signature extracts information about the statistical moments. More details can be found in~\cref{app:moments}. This, paired with the signature's summarization and approximation capabilities \citep{Chevyrev:16}, makes it a natural candidate for path processing.
        The $k$-dimensional randomized signature
        $S_R(G)$ of $G$ is the vector $\bm{z}_T \in \dR^{k}$
        which is obtained via $d$ incremental updates.
        The randomized signature matrix of the graph $\bm{Z} \in \dR^{d \times k}$ refers to the collection of $d$ update steps leading to $\bm{z}_T$.
        By interpreting a graph with $N$ nodes as a path with $N$ coordinate paths as illustrated in \Cref{fig:graph_as_path}, we can process global feature information of the entire graph in each iterative step of 
        \Cref{alg:randomized}. Consequently, we have interpreted node information of any graph structured data as a path,
        and have shown how to calculate the randomized signature thereof. 
        We 
        define a 
        procedure for edge embedding.

        \textbf{Edge embedding. }
        Following~\citet{Roth:03, Duda:01, Schoelkopf:98}, we obtain an edge embedding
        via the eigendecomposition of the adjacency matrix $\bm{A} \in \dR^{N \times N}$:
        \begin{align}\label{eq:escoring}
            -\frac{1}{2}\bm{Q}\bm{A}\bm{Q} = \bm{V}\boldsymbol{\Lambda}\bm{V}^{T}\ ,
        \end{align}
        where $\displaystyle \bm{Q} = \bm{I}_{N} - \nicefrac{1}{N}\left(1_{1}, \ldots{}, 1_{N}\right)^{T}\left(1_{1}, \ldots{}, 1_{N}\right)$ is used for normalization, and $\bm{I}_N$ is the $N \times N$ identity matrix,  
        $\bm{V} = \left({\bm v}_{1}, \ldots{}, {\bm v}_{N}\right)$ is a matrix with the eigenvectors as its columns, and $\boldsymbol{\Lambda} = \diag\left(\lambda_{1}, \ldots{}, \lambda_{N}\right)$ is the diagonal matrix of the accompanying eigenvalues, sorted in decreasing order. The edge embedding matrix $\bm{E}_{\bm{A}} \in \dR^{N \times m}$ is obtained by 
        \begin{align}\label{eq:embeddings}
            \bm{E}_{\bm{A}} = \bm{V}_{m}\sqrt{\boldsymbol{\Lambda}_{m}} \ ,
        \end{align}
        where $0< m < N$ is the embedding dimension (hyperparameter), $\boldsymbol{\Lambda}_{m} \in \dR^{m \times m}$ is the submatrix with the $m$ largest eigenvalues on its diagonal and $\bm{V}_{m} \in \dR^{N \times m}$ with the corresponding eigenvectors. For more details about the embedding procedure see \Cref{app:emb}.
        Similar to the node information, we can now view $\bm{E}_{\bm{A}}^{T} \in \dR^{m \times N}$ as $N$ coordinate paths sampled at $m$ points,
        and thus have also found a way to interpret edge information as a path. The edge embedding procedure also holds if $\bm{A}$ is replaced with an arbitrary dissimilarity matrix, e.g., when positional information is given as with point clouds, which is exemplified in \cref{fig:coma_faces}.
        
        \section{G-Signatures}\label{sec:architecture}
        We follow the Encode-Process-Decode framework of~\citet{Battaglia:18} and~\citet{SanchezGonzalez:20}. 
        For the encoding, we introduce \emph{graph conversion} which allows us to ``convert'' graph structured data to a latent representation. In
        latent space, we process the 
        paths via randomized signature layers which traverse the converted graph on a per-feature basis 
        by introducing
        the concept of \emph{latent space path mappings} (LSPM).
        Finally, we decode the extracted information into a target representation.
        
        \textbf{Graph conversion. }
        For a given graph $\cG = (\cV,\cE)$ with $N$ nodes and node features $\bm{h}_i \in \dR^{d}$
        attached to them, we can summarize the nodes to $\bm{G} \in \dR^{d \times N}$, and the transposed edge embedding of \Cref{eq:embeddings} to
        $\bm{E}_{\bm{A}}^T \in \dR^{m \times N}$.
        Depending on the task at hand, we can either use $\bm{G}$, 
        or $\bm{E}_{\bm{A}}^T$, or concatenate them in the first dimension.
        For the encoding, we sequentially map each dimension into a latent space such that we get a latent graph representation $\bm{X} \in \dR^{h_1 \times h_2}$ with hidden dimensions $h_1$ and $h_2$. This representation can be interpreted as a path $X:[0,T] \rightarrow \dR^{h_2}$ sampled at $h_1$ points $1 < \ldots < h_1 = T$.
        The current graph conversion procedure is designed for homogeneous graphs, 
        i.e., graphs with the same number of nodes and the same connectivity. 
        We think that the graph conversion concept can be
        extended to heterogeneous graphs.
        Similarly, the edge embedding can be extended to edge features beyond scalar connectivity information. 
        We will address these two extensions in future work.
        Both, graph conversion and our edge embedding strategy allow us to efficiently compress large-scale graphs without losing edge connectivity information. This in turn forms the basis for efficient propagation of graph information as evidenced in experiments where our method needs only a fraction of memory and runtime compared to baselines.
        
        \textbf{Latent space path mappings (LSPM). }
        LSPM comprises a sequence of stacked randomized signature layers.
        More precisely, for $l \in \{1, \dots, L\}$ where $L$ is the number of signature layers the input path to the $l$\textsuperscript{th} randomized signature layer $\bm{X}^l \in \dR^{h_1 \times h_2}$ is mapped to an output path $\bm{X}^{l+1} \in \dR^{h_1 \times h_2}$ in the same space. The mapping is based on the procedure as described in \cref{alg:randomized},
        with the difference that $\bm{A_i}$,  and $\Bb_i$ are learnable parameters.
        G-Signatures learn how to refine the random mappings via gradient descent, that is,
        they learn to extract task-relevant information from the signature representation of a path.
        This is done in a bidirectional manner along 
        the axis that represents the time steps of the underlying path 
        $1 < \ldots < h_1 = T$,
        resulting in the randomized signature matrix $\bm{Z}^l \in \dR^{h_1 \times 2 k}$ 
        (again, $\bm{Z^l}$ is a path itself).
        We map the 
        randomized signature matrix $\bm{Z}^l \in \dR^{h_1 \times 2 k}$ back to $\bm{X}' \in \dR^{h_1 \times h_2}$,
        adding it to the input path via a residual connection: $\bm{X}^{l+1} = \bm{X}^l + \bm{X}'$. Consequently, we can stack several randomized signature layers, and thus enable mappings between paths in latent space.
        The randomized signature layers are easily parallelizable to multiple signature ``heads'' in similar vein to \mbox{Multihead Attention~\citep{Vaswani:17}.} An overview of the LSPM procedure is visualized in the Appendix in~\cref{fig:lspm_illustrated} with a more detailed sketch in~\cref{fig:architecture_overview}.
        
        \textbf{Characteristics of LSPM implementation.}
        Since 
        the naive application of the randomized signature does not produce desirable results
        we made important adjustments. 
        In order to obtain an adequate learning behavior, three important adjustments to ~\cref{alg:randomized} are required: 
        (i) \textbf{sparsity} of the learnable weights $\bm{A}_{i}$,
        (ii) suitable weight \textbf{initialization}, and
        (iii) proper \textbf{activation} functions.
        Concerning (i), we experimentally found that dense parameters $\bm{A}_i$
        dilute the signal $\bm{X}^l$ (similar to graph over-smoothing) leading to outputs with high noise-levels.
        We prevent the signal from diluting by sparsifying $\bm{A}_i$.
        Furthermore, we found that initializing the components of $\bm{A}_i, \Bb_i$ from $\mathcal{N}(0, 1)$ is prone to produce high weight values that lead to unfavorable learning dynamics.
        We, thus, reduce the weight magnitude by adaptively reducing the variance depending on the signature size $k$ to $\mathcal{N} \left( 0, \nicefrac{1}{k} \right)$.
        Finally, to prevent exploding signal values,
        we use the identity as activation 
        function scaled by $\nicefrac{1}{h_{2}}$.
        Exploding signal values arise due to the double summation of the bidirectional signature
        for large $h_{2}$ and non-vanishing $X_j^i$ in \Cref{alg:randomized}.
        The activation function compensates for large numbers of summation steps in case of a high-dimensional path space and implicitly reduces the variance of $\bm{A}_i, \Bb_i$ by a factor of $\nicefrac{1}{h_{2}^2}$. 
        We validate the importance of these adjustments by ablation studies shown in~\Cref{app:ablat}.
        
        \textbf{Decoding.}
        For the decoding, we use a similar approach as for the encoding, and sequentially map the latent space dimensions to the output dimensions.
                
        \section{Experiments}\label{sec:experiments}
        We test G-Signatures on several tasks, i.e.,
        (i) global graph classification on the CoMA dataset~\citep{Ranjan:18}, (ii) node regression on the Kardar–Parisi–Zhang (KPZ)~\citep{Kardar:86} equation with varying degrees of noise, and (iii) large graph edge regression on estimated time of arrival (ETA) tasks with varying sparsity degrees. 
        For non-gridded graph structured data, i.e., CoMA and ETA tasks, we compare against Gated Graph Convolutional Networks (GGCNs)~\citep{Bresson:17} and Graph Transformers (GTs)~\citep{Dwivedi:21} since they are amongst the best performing methods in several long range graph benchmarks~\citep{Dwivedi:22}.
        GGCNs~\citep{Bresson:17} are a GCN \citep{Kipf:17} based representative that uses learned edge gates to improve the aggregation procedure.
        GTs~\citep{Dwivedi:21} generalize transformer networks on arbitrary graphs via node attention and pairwise modification of the attention scores via edge attributes.
        For the KPZ equation, we compare against ResNets~\citep{He:16} and Fourier Neural Operators (FNOs)~\citep{Li:20fno} on regularly-gridded data.
        We describe each task in detail, further information can be found in \Cref{app:hyperparameters}.        
        
        \subsection{Recognizing facial expressions}
\begin{figure}[ht]
            \begin{center}
            \hfill
            \begin{subfigure}[b]{0.24\columnwidth}
                \begin{center}
                \includegraphics[width=\columnwidth]{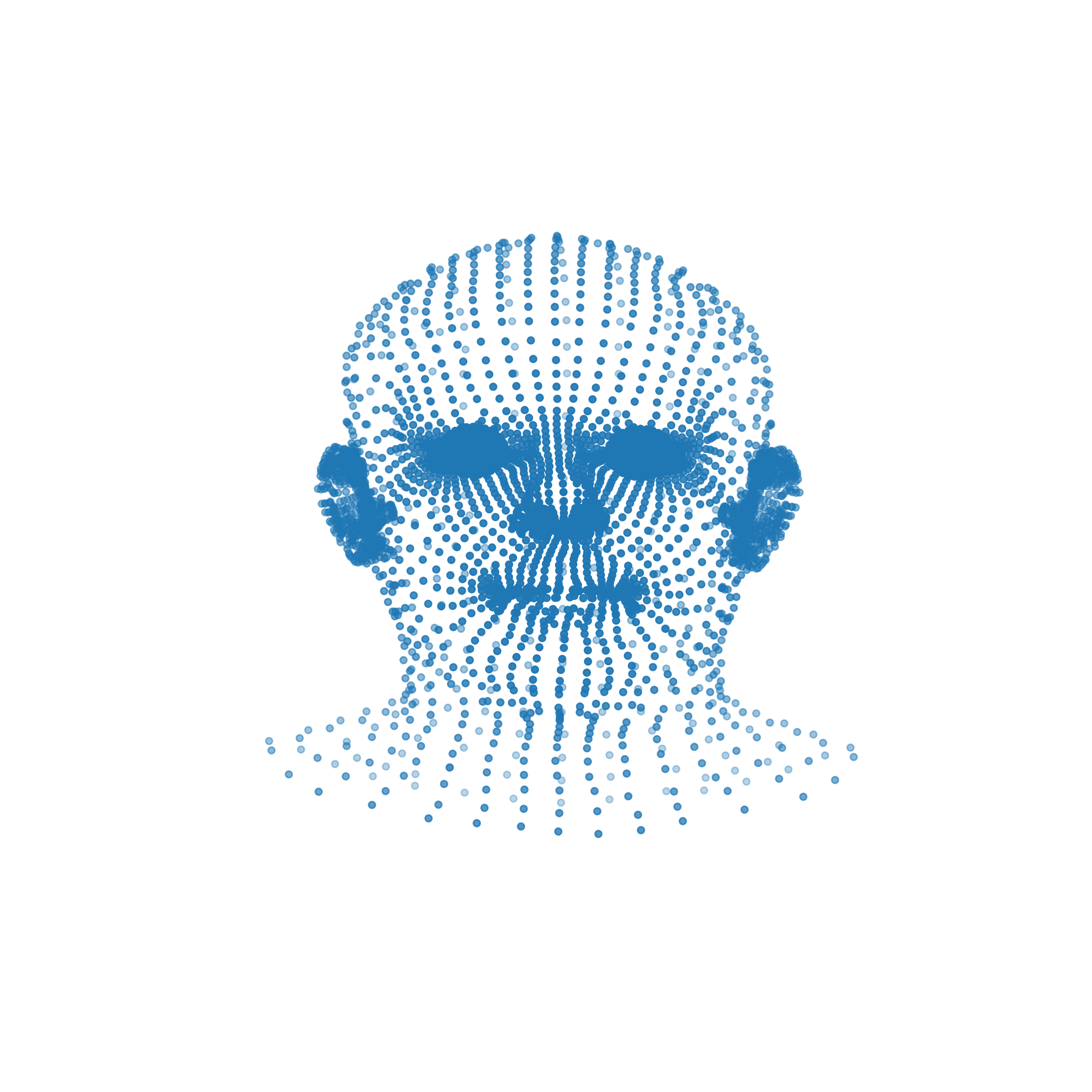}
                \end{center}
            \end{subfigure}
            \hfill
            \begin{subfigure}[b]{0.24\columnwidth}
                \begin{center}
                \includegraphics[width=\columnwidth]{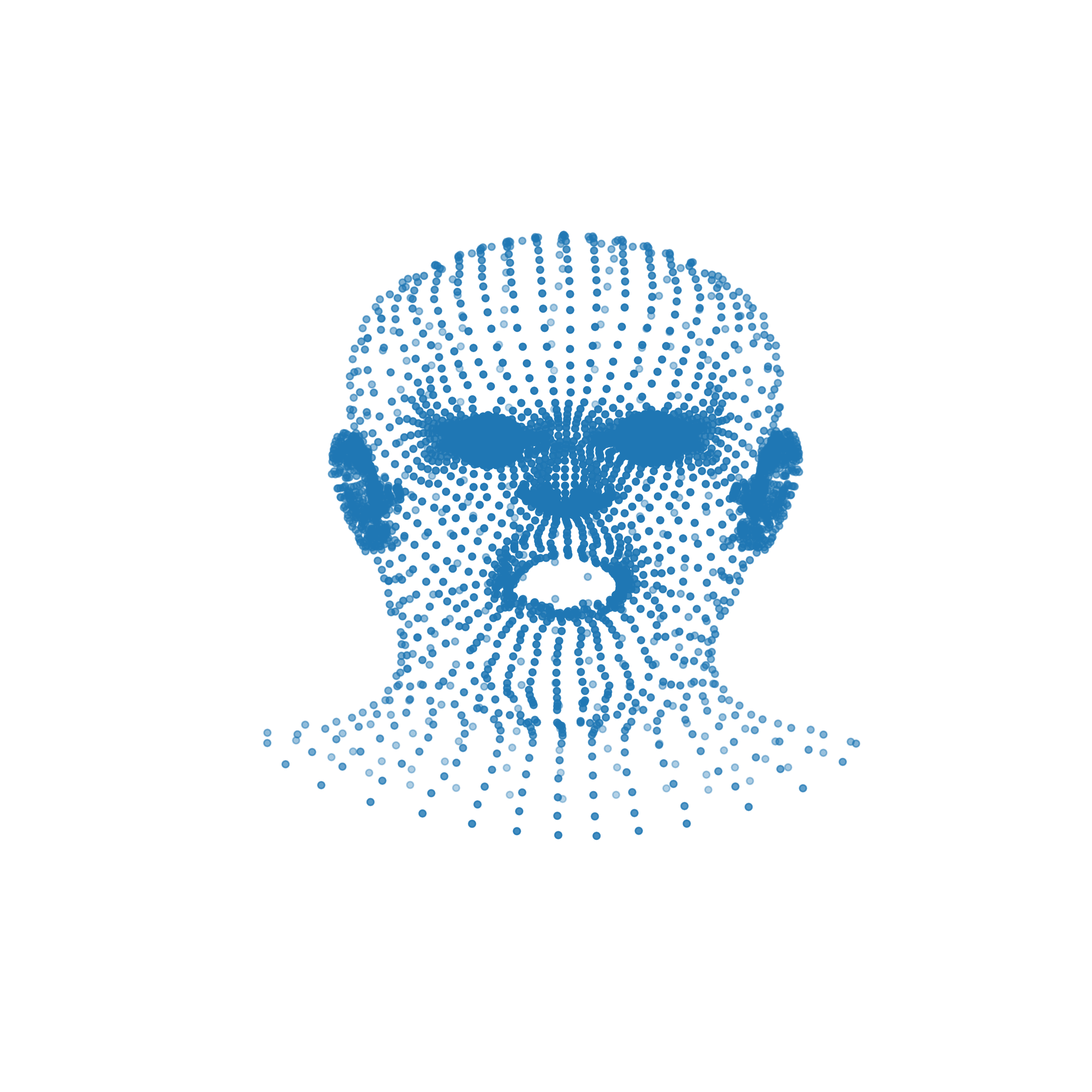}
                \end{center}
            \end{subfigure}
            \hfill
            \begin{subfigure}[b]{0.24\columnwidth}
                \begin{center}
                \includegraphics[width=\columnwidth]{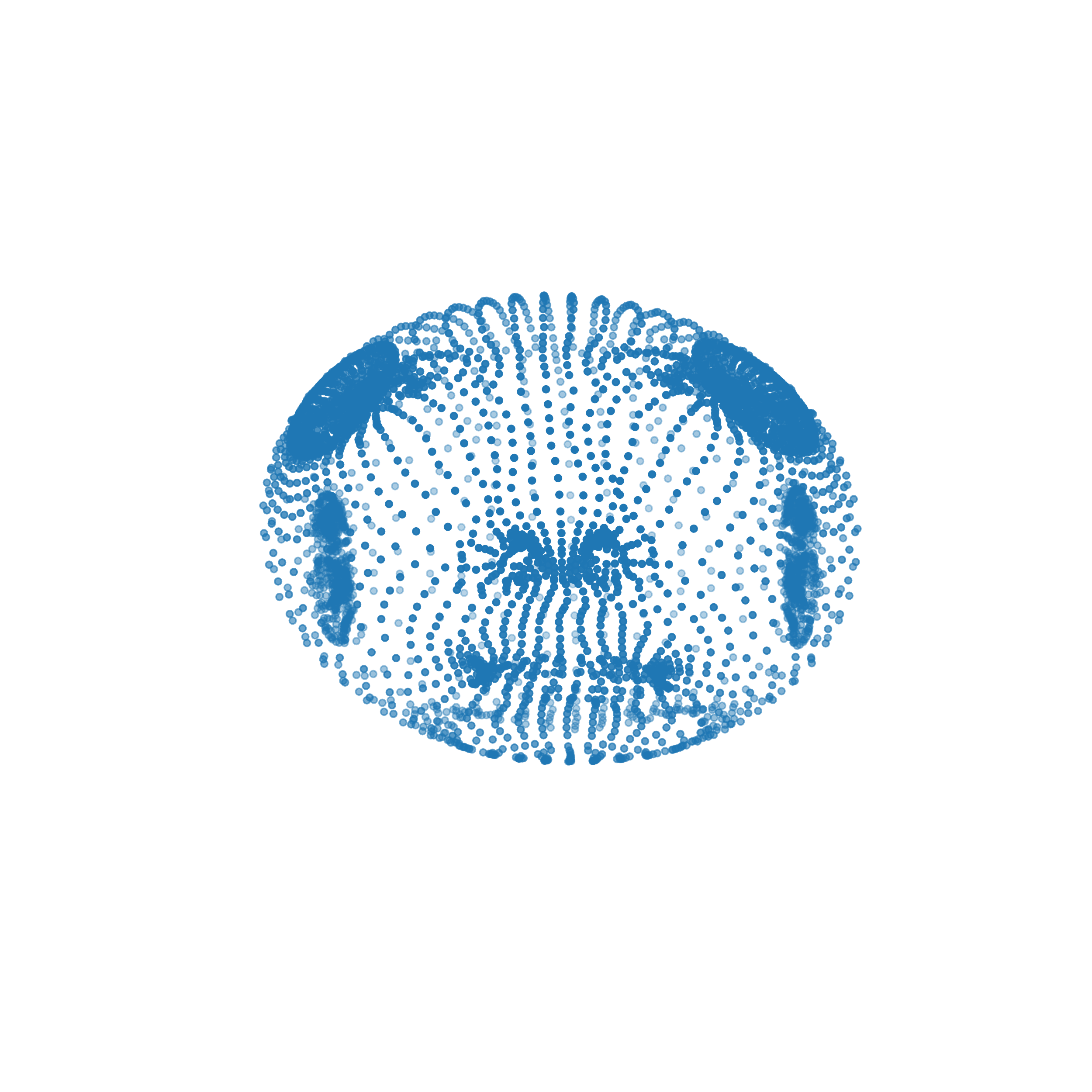}
                \end{center}
            \end{subfigure}
            \hfill
            \begin{subfigure}[b]{0.24\columnwidth}
                \begin{center}
                \includegraphics[width=\columnwidth]{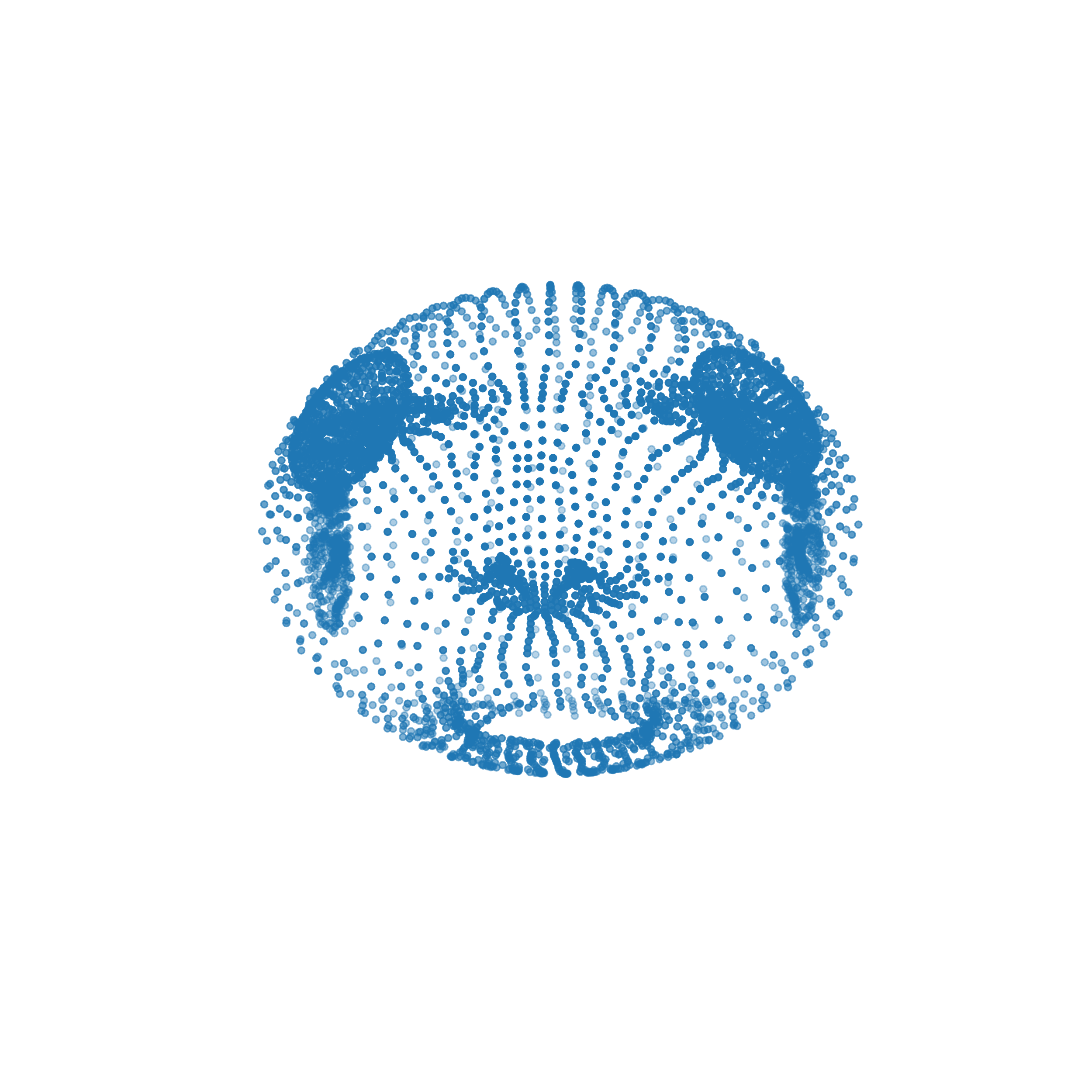}
                \end{center}
            \end{subfigure}
            \hfill
            \caption{
            Exemplary samples in the CoMA dataset. Left: the original datapoints, right: their corresponding embedded versions, according to \citep{Roth:03}.
            Most relevant information of the original graph is visually contained in the converted graph representation.}\label{fig:coma_faces}
            \end{center}
        \end{figure}

        The convolutional mesh autoencode (CoMA) dataset~\citep{Ranjan:18} consists of 20465 graphs where each graph comprises 5023 nodes with 3 (positional) node features each, and 29990 edges, see \cref{fig:coma_faces} for visualized datapoints. 
        The task is to predict the facial expression for 12 different target labels. 
        We use the 1483 samples test dataset as provided by the CoMA dataset implementation, and split the given training dataset into 17499 training and 1483 validation samples in a stratified way.         
        For all methods, i.e., G-Signatures, GGCNs, and GTs, we use the edge embedding of \Cref{eq:embeddings} and set $m=3$. This results in network inputs of $\dR^{3 \times 5023}$.
        GGCNs use 4 layers with a hidden dimension of $70$, resulting overall in $104$k parameters. GTs use $10$
        attention layers, with $8$ attention heads each and a hidden dimension of $80$, resulting in $913$k parameters.
        For GGCNs, adjacency information is additionally provided to the graph convolution layers.
        G-Signatures benefit from the fact that we convert the graph to a latent space path of dimension $h_1 \times h_2 = 39\times{}39$, which substantially reduces the cost of memory and compute.
        G-Signatures outperform the baseline methods with an accuracy (evaluated on three replicates) of $93.74 \pm 0.76$, compared to $92.85 \pm 0.42$ for GTs, and $76.74 \pm 1.19$ for GGCNs. Most notably, G-Signatures are much less memory consumptive, and have much lower inference and training times as shown in \Cref{tab:coma_results}.
        
         \begin{table*}[ht]
             \begin{center}
                 \begin{tabular}{l S[table-format=2.2(2),separate-uncertainty] S[table-format=3] S[table-format=2] S[table-format=4] S[table-format=1.3] S[table-format=1.3]}
                 \toprule
               {\makecell{\bf Method}} & {\makecell{\bf Accuracy}} & {\makecell{\bf \#Parameters[k]}} & {\makecell{\bf \#Layers}} & {\makecell{\bf Memory[MiB]}} & {\makecell{\bf Fwd.[s]}} & {\makecell{\bf Fwd.+Bwd.[s]}} \\
                 \midrule
                 GGCNs & 76.74(119) & 104 & 4 & 1581 & 0.005 & 0.025 \\
                 GTs & 92.85(42) & 913 & 10 & 1875 & 0.052 & 0.101 \\
                 G-Sigs. & 93.74(76) & 430 & 1 & 1361 & 0.002 & 0.022 \\
                 \bottomrule
                 \end{tabular}
             \end{center}
             \caption{Comparison of accuracy, parameter count, network depth, and memory/time consumption of compared methods on the CoMA dataset. Memory and time consumption is measured per datapoint.}\label{tab:coma_results}
         \end{table*}
        
        \subsection{Surrogate models for stochastic PDEs}
        Next, we show that our method can competitively perform with strong baselines on tasks it was not primarily designed for, making it a candidate for more general usage scenarios.
        More precisely, we stress-test the ability of G-Signatures to model spatially connected data, and, thus, force G-Signatures to not only traverse information on a per-feature basis, but also aggregate across nodes. Furthermore, using regularly gridded data, and comparing against state-of-the-art methods on those, i.e., FNOs~\citep{Liu:20} and ResNets~\citep{He:16}, presents a litmus test for each graph-tailored method.
        We, therefore, aim to learn surrogates on temporal evolving noisy data, 
        that is from data obtained from numerical partial differential equation (PDE) solvers.
        Concretely, we look at the Kardar–Parisi–Zhang (KPZ) equation,
        which is a stochastic PDE that describes
        the spatial temporal change of $\bm{u}(x,t)$:
        \begin{align}
            \frac{\partial \bm{u}}{\partial t} = \nu \Delta \bm{u} + \frac{\lambda}{2} (\nabla \bm{u})^2 + \eta(x,t)
            \label{eq:kpz}
        \end{align}
        where $\nu$ and $\lambda \in \dR$ are diffusion and viscosity coefficient, $\Delta$ and $\nabla$ are Laplace- and Nabla-operator, and $\eta(x,t)$ is white Gaussian noise, find more information in \cref{app:pde}.
        The task can be interpreted as node regression task on an equidistant graph where the temporal inputs are the node features.                 
        We test for different noise levels, comparing against FNOs~\citep{Li:20fno}, and ResNets~\citep{He:16}, following the experimental setup presented in \citet{Brandstetter:22b}.
        Both architectures are built from translation equivariant layers, and are heavily used on equidistant grids.
        The input and output of the models are $10$ timesteps in the channel dimension. The ResNet18 is a 1D variant of~\citet{He:16} with $4\times2$ residual blocks,
        and 128 hidden channels. Overall this results in 801k parameters.
        The FNO architecture uses 4 1D FNO layers as proposed in \citet{Li:20fno} with 12 Fourier modes,
        resulting in 863k parameters.
        These layers have residual connections and are intertwined with GeLU~\citep{Hendrycks:16} non-linearities.
        G-Signatures convert the graph to a latent space representation $\bm{X} \in \dR^{h_{1} \times h_{2}}$
        with $81 \leq h_{1} = h_{2} \leq 89$.
        We use one randomized signature layer for LSPM, overall resulting in an architecture with 53.7k parameters for the high noise dataset, 101k parameters for the low noise dataset, and 64.4k parameters for the zero noise dataset.
        \newcommand{\SIKPZ}[2]{{\multirow{1}{*}{\SI[tight-spacing=false]{#1}}$\pm$ {\multirow{1}{*}{\SI[tight-spacing=false]{#2}}}}}
        \begin{table*}[ht]
            \begin{center}
                \begin{tabular}{
                    l
                    S[table-format=1e-1]
                    c
                    c
                    c}
                \toprule
                    {\makecell{\bf Noise\\\bf level}} &
                    {\makecell{\bf Standard\\\bf deviation}} &
                    {\makecell{\bf ResNets[MSE]}} &
                    {\makecell{\bf FNOs[MSE]}} &
                    {\makecell{\bf G-Signatures[MSE]}} \\
                \midrule
                     High   & 5e-3 & \SIKPZ{4.501e-3}{0.004e-3} & \SIKPZ{4.866e-3}{0.022e-3} & \SIKPZ{4.774e-3}{0.013e-3} \\
                     Low    & 1e-3 & \SIKPZ{0.991e-3}{0.018e-3} & \SIKPZ{1.197e-3}{0.025e-3} & \SIKPZ{0.980e-3}{0.007e-3} \\
                     Zero   & 0e-3 & \SIKPZ{0.075e-3}{0.006e-3} & \SIKPZ{0.160e-3}{0.014e-3} & \SIKPZ{0.068e-3}{0.002e-3} \\
                \bottomrule
                \end{tabular}
            \end{center}
            \caption{
            Performance 
            on the KPZ datasets with three different noise levels. The reported deviation is the standard error of the mean. The noise level is steered by the standard deviation of Gaussian noise $\eta$. G-Signatures benefit from lower noise levels.}\label{tab:results_kpz}
        \end{table*}
        In \Cref{tab:results_kpz} model performances are compared at different noise levels. See \Cref{fig:results_kpz} of the Appendix for more information.
        Although not tailored to regularly gridded data,
        G-Signatures are competitive with FNO and ResNet baselines, surpassing both baselines in the zero, and low noise settings. In the high noise setting, G-Signatures perform second best.
        This shows that G-Signatures keep up with strong baselines on regularly gridded tasks 
        without using convolution operations,
        making it a more generally applicable method not limited to typical graph representation learning tasks. 

        \subsection{Estimated time of arrival}
        The estimated time of arrival (ETA) is the time it is expected to take a vehicle, or a person, to arrive at a certain place~\citep{Derrow_Pinion:21,Hu:22,Wang:18}.
        We model the ETA task by a graph with adjacency matrix $\bm{A}$. Each component $A_{ij}$ represents the time it takes to get from node $i$ to node $j$ if the nodes are connected.
        On a map the nodes could for example represent intersections and only if a connecting street between nodes $i$ and $j$ exists, the value in the adjacency matrix will be the time it takes to get from $i$ to $j$. 
        Otherwise, we set the value of the edge connecting node $i$ and $j$ to infinity. The task is to predict the shortest travel time from node $i$ to $j$ by possibly traversing other nodes. By increasing the sparsity in the adjacency matrix, more nodes have to be traversed to get from node $i$ to node $j$. 
        We conducted experiments on graphs with 3 different node counts with 3 different connectivity levels (number of edges) each, resulting in 9 experimental settings, find more information in \cref{app:eta}.
        Gated GCNs and Graph Transformers additionally use connectivity information about the nodes as edge features for information aggregation across nodes.
        GGCNs use $10$ layers with a hidden dimension of $70$, resulting overall in $254$k parameters. GTs use $8$
        attention layers, with $8$ attention heads each and a hidden dimension of $32$, resulting in $119$k parameters.
        G-Signatures again benefit from the fact that we convert the graph to a latent space representation $\bm{X} \in \dR^{h_{1} \times h_{2}}$ 
        with $32 \leq h_{1} = h_{2} \leq 95$.
        This substantially reduces the cost of memory and compute.
        The more the connectivity decreases the more important global aspects of the graph become since connecting two nodes can require propagating over many nodes in the graph. 
        Each node-pair needs to be regressed as a scalar and we use MSE to measure the performance.
        \begin{table}[ht]
                \begin{tabular}{S[table-format=4.0]S[table-format=1.2e1]S[table-format=5.0]S[table-format=5.0]S[table-format=5.0]}
                \toprule
                {\makecell{\bf{}\#{}Nodes}} & {\makecell{\bf \#{}Edges}} & {\makecell{\bf GGCNs}} & {\makecell{\bf GTs}} & {\makecell{\bf G-Sigs.}}\\
                \midrule
                          & 2.50e+5 &  4401 &  4881 &  1373 \\
                      500 & 1.25e+5 &  2901 &  3205 &  1383 \\
                          & 2.50e+4 &  1815 &  1823 &  1353 \\
                \cmidrule{1-3}\cmidrule{4-5}
                          & 1.00e+6 & 13257 & 14861 &  1425 \\
                     1000 & 5.00e+5 &  7355 &  9399 &  1417 \\
                          & 1.00e+5 &  2655 &  2987 &  1411 \\
                \cmidrule{1-3}\cmidrule{4-5}
                          & 4.00e+6 & 48905 & 55023 &  1655 \\
                     2000 & 2.00e+6 & 39756 & 46165 &  1635 \\
                          & 4.00e+5 &  6295 &  7027 &  1639 \\
                \bottomrule
                \end{tabular}
            \caption{
                Memory consumption of compared methods for processing one graph during training on all estimated time of arrival (ETA) datasets measured in MiB.}\label{fig:memory_eta}
        \end{table}
        \begin{figure}[ht]
            \includegraphics[width=\columnwidth]{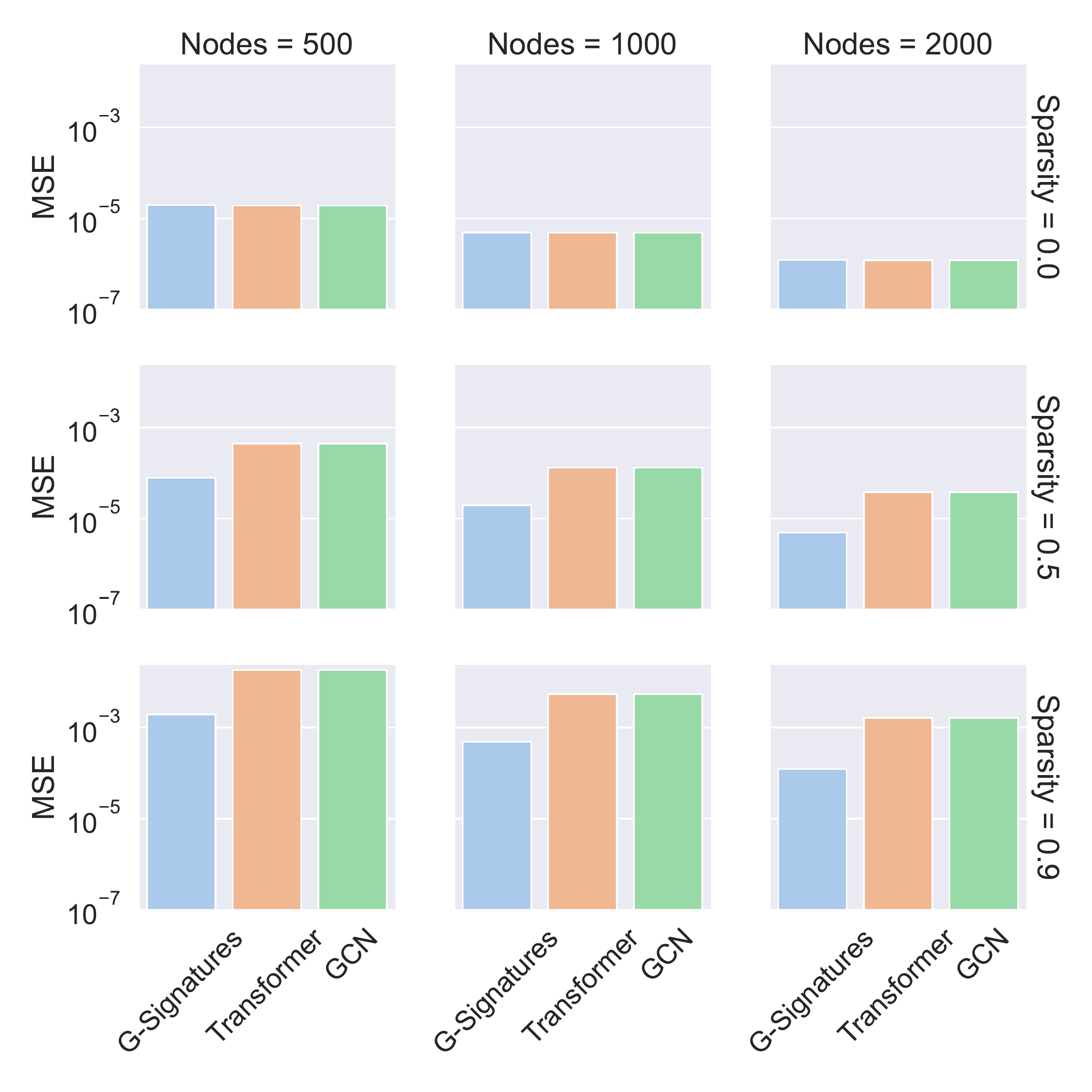}
            \caption{
                Results on
                ETA. 
                When increasing the sparsity of the graphs, G-Signatures substantially outperform the competitors, often by an order of magnitude. See~\cref{tab:results_eta_mse} of the Appendix for precise numerical results with error bounds.}\label{fig:results_eta}
        \end{figure}
        Results are reported in~\Cref{fig:results_eta} and~\cref{tab:results_eta_mse} in Appendix,
        confirming the advantage of global graph propagation via G-Signatures. All compared methods have very similar performance in case of high connectivity. This attributes to the fact that in these scenarios the propagation of global information is less important since all node pairs share an edge thus sufficient information can be processed locally.
        For example, GGCNs need relatively few steps to propagate information through the graph.
        However, when increasing the sparsity of the graphs, G-Signatures substantially outperform the competitors, often by an order of magnitude.
        GTs, to exchange information, are not restricted to the local neighborhood of each node at a given message passing iteration.
        Nevertheless, for ETA prediction information collected along the possible paths between distant source and sink nodes has to be propagated over many transit nodes what also exposes GTs to the risk of over-squashing.
        Whereas G-Signatures can easily propagate information on a global scale by using only a fraction of the memory consumption compared to state-of-the-art methods, summarized in~\Cref{fig:memory_eta}.        
        \section{Conclusion}
        We have introduced, G-Signatures, a novel learning paradigm for learning on graph structured data. G-Signatures are built around the concepts of graph conversion, which allows us to treat graph structured data as paths in latent space, and latent space path mapping (LSPM), which performs global graph propagation via randomized signature layers.
        Consequently, in contrast to conventional GNNs or Graph Transformers, G-Signatures excel at extracting global graph properties with substantially reduced memory and compute budget.
        We have further shown that G-Signatures are a more generally applicable method not limited to GNN-specific tasks.
        
        \textbf{Limitations and future work.}
        In the current setting, G-Signatures are designed for homogeneous graphs, i.e., graphs with the same number of nodes and the same connectivity. For future work, we aim to extend G-Signatures towards heterogeneous graphs to provide applicability for a larger set of problems, e.g., molecular modeling.
        Furthermore, we plan to extend the edge embedding to edge features beyond scalar connectivity information.
        Finally, 
        the current
        version
        traverses paths
        along one dimension, but 
        we
        aim to traverse paths along concurrent dimensions, and thus combine information with a much denser and richer content.

        \bibliography{gsignatures}
        
        \newpage
        \appendix
        \onecolumn

        \setcounter{secnumdepth}{1}

        \section{Statistical moments from the signature} \label{app:moments}
        Interpreting the path $X$ as random variable with a realization resulting in the dataset $\{X_i\}_{i=1}^N$ \cite{Chevyrev:16} have shown that truncating the signature at level $L$ determines statistical moments up to level $L$. This is exemplary demonstrated for the first two moments. Using the canonical cumulative lead-lag embedding of $\{X_i\}_{i=1}^N$ we get
        \begin{align*}
            S(X)^1 = S(X)^2 &= \sum_{i=1}^N X_i \\
            S(X)^{1,1} = S(X)^{2,2} &= \frac{1}{2} \left( \sum_{i=1}^N X_i \right)^2 \\
            S(X)^{1,2} &= \frac{1}{2} \left[ \left( \sum_{i=1}^N X_i \right)^2 + \sum_{i=1}^N X_i^2 \right] \\
            S(X)^{2,1} &= \frac{1}{2} \left[ \left( \sum_{i=1}^N X_i \right)^2 - \sum_{i=1}^N X_i^2 \right].
        \end{align*}
        From this we get that
        \begin{align*}
            Mean(X) &= \frac{1}{N} S(X)^1 \\
            Var(X) &= - \frac{N+1}{N^2} S(X)^{1,2} + \frac{N - 1}{N^2} S(X)^{2,1}.
        \end{align*}
        For a 1-dimensional path $X$ its corresponding signature is
        \begin{align*}
            S(X)= \left( 1, X_N - X_1, \frac{(X_N - X_1)^2}{2!}, \cdots, \frac{(X_N - X_1)^k}{k!}, \cdots \right).
        \end{align*}
        Similarly, interpreting $X$ as stochastic process its expected signature is then
        \begin{align*}
            \mathbb{E}[S(X)] = \left( 1, \mathbb{E}[X_N - X_1], \frac{\mathbb{E}[(X_N - X_1)^2]}{2!}, \cdots, \frac{\mathbb{E}[(X_N - X_1)^k]}{k!}, \cdots \right)
        \end{align*}
        which motivates the question of how this moment-like sequence relates to the law of $X$. \cite{Chevyrev:18} have shown that under certain assumptions the expected signature of a path $X: [a,b] \rightarrow \mathbb{R}^d$ indeed characterizes its law such that it can be thought of as a generalization of the moment-generating function of a real-valued random variable. 
        
        \section{Edge Embedding} \label{app:emb}
        Following~\citet{Roth:03, Duda:01, Schoelkopf:98}, edge information is included into the nodes 
        via the eigendecomposition of a dissimilarity matrix $\bm{D} \in \dR^{N \times N}$:
        \begin{equation}\label{eq:escoring_appendix}
            \begin{gathered}
            \bm{Q} = \bm{I}_{n} - \frac{1}{n}\left(1_{1}, \ldots{}, 1_{n}\right)^{T}\left(1_{1}, \ldots{}, 1_{n}\right)\ , \\
            \bm{S}^c = -\frac{1}{2}\bm{QDQ} = \bm{V\Lambda{}V}^{T}\ ,
            \end{gathered}
        \end{equation}
        This formula is derived from the following statements.
        Let $\bm{M} \in \{0,1\}^{n \times k}$ be a binary stochastic assignment matrix  with $\sum_{\nu=1}^k M_{i\nu} = 1$.
        The pairwise clustering cost function is used to derive meaningful embedding vectors $\{\bm x_{i}\}_{i=1}^{n}$ from a corresponding dissimilarity matrix $\bm{D}$ with \mbox{$D_{ij} = d(\bm{x}_{i}, \bm{x}_{j})$.} The pairwise clustering cost function is defined as:
        \begin{align}\label{eq:hpc}
            H^{pc} =& \frac{1}{2}\sum_{\nu{}=1}^{k}{\frac{\sum_{i=1}^{n}{\sum_{j=1}^{n}{M_{i\nu{}}M_{j\nu{}}D_{ij}}}}{\sum_{l=1}^{n}{M_{l\nu{}}}}}\ ,
        \end{align}
        where $\bm{D}$ is an arbitrary dissimilarity matrix between embedding vectors $\{\bm x_{i}\}_{i=1}^{n}$. Assuming $d({\bm x}, {\bm y}) = \|{\bm x} - {\bm y}\|^{2}$ as the dissimilarity function of choice the pairwise clustering cost function $H^{pc}$ is equal to the $k$-means cost function iff $\bm{S}^{c}$ as described in \Cref{eq:escoring_appendix} is positive semidefinite~\citep{Duda:01,Roth:03}.
        Minimizing $H^{pc}$ leads to an optimal assignment matrix $\bm{M}$.
         For possibly indefinite $\bm{S}^{c}$, the given dissimilarity matrix $\bm{D}$ needs to be centralized and shifted.
        A diagonal shift of \cref{eq:escoring_appendix} corresponds to an off-diagonal shift of the dissimilarity matrix $\bm{D}$ with the smallest eigenvalue of $\lambda(\bm{S}^c)$:
        \begin{align}
           \Tilde{\bm{D}} = \bm{D} - 2\lambda{}_{n}\left(\bm{S}^{c}\right)\left({\bm e}_{n}^T {\bm e}_{n} - \bm{I}_{n}\right)\ .
        \end{align}
        Substituting $D_{ij}$ with $\Tilde{D}_{ij}$ in \cref{eq:hpc} gives us the $k$-means cost function.
        Equivalently to \cref{eq:escoring_appendix}, the centralized squared Euclidean scoring matrix is
        \begin{align}\label{eq:escoring_squared}
           \bm{\Tilde{S}}^{c} = -\frac{1}{2}\bm{\Tilde{D}}^{c} = -\frac{1}{2}\bm{Q\Tilde{D}Q}\ .
        \end{align}
        Following the definition of the kernel principal component analysis (PCA), the original embedding vectors can be reconstructed by an eigendecomposition of \cref{eq:escoring_decomposition}~\citep{Schoelkopf:98,Roth:03}:
        \begin{equation}\label{eq:escoring_decomposition}
            \begin{gathered}
            \bm{Q} = \bm{I}_{n} - \frac{1}{n}\left(1_{1}, \ldots{}, 1_{n}\right)^{T}\left(1_{1}, \ldots{}, 1_{n}\right)\ , \\
            \bm{S} = -\frac{1}{2}\bm{QDQ} = \bm{V\Lambda{}V^{T}}\ ,
            \end{gathered}
        \end{equation}
        where $\bm{V} = \left({\bm v}_{1}, \ldots{}, {\bm v}_{n}\right)$ is a matrix with the eigenvectors as its columns, and $\bm{\Lambda} = \diag\left(\lambda_{1}, \ldots{}, \lambda_{n}\right) $ is the diagonal matrix of the corresponding eigenvalues, sorted in decreasing order. The embedding vectors are recovered by
        \begin{align}\label{eq:embeddings_appendix}
            \bm{X} = \bm{V}_{m}\left(\bm{\Lambda_{m}}\right)^{0.5}\ ,
        \end{align}
        where $0< m < n$ is the embedding dimension, $\bm{X}\in{}\mathbb{R}^{n\times{}m}$,  $\boldsymbol{\Lambda}_{m} \in \dR^{m \times m}$ is the submatrix with the $m$ largest eigenvalues on its diagonal and $\bm{V}_{m} \in \dR^{N \times m}$ with the corresponding eigenvectors.
        
        \begin{figure}[ht]
            \begin{center}
            \includegraphics[width=\textwidth]{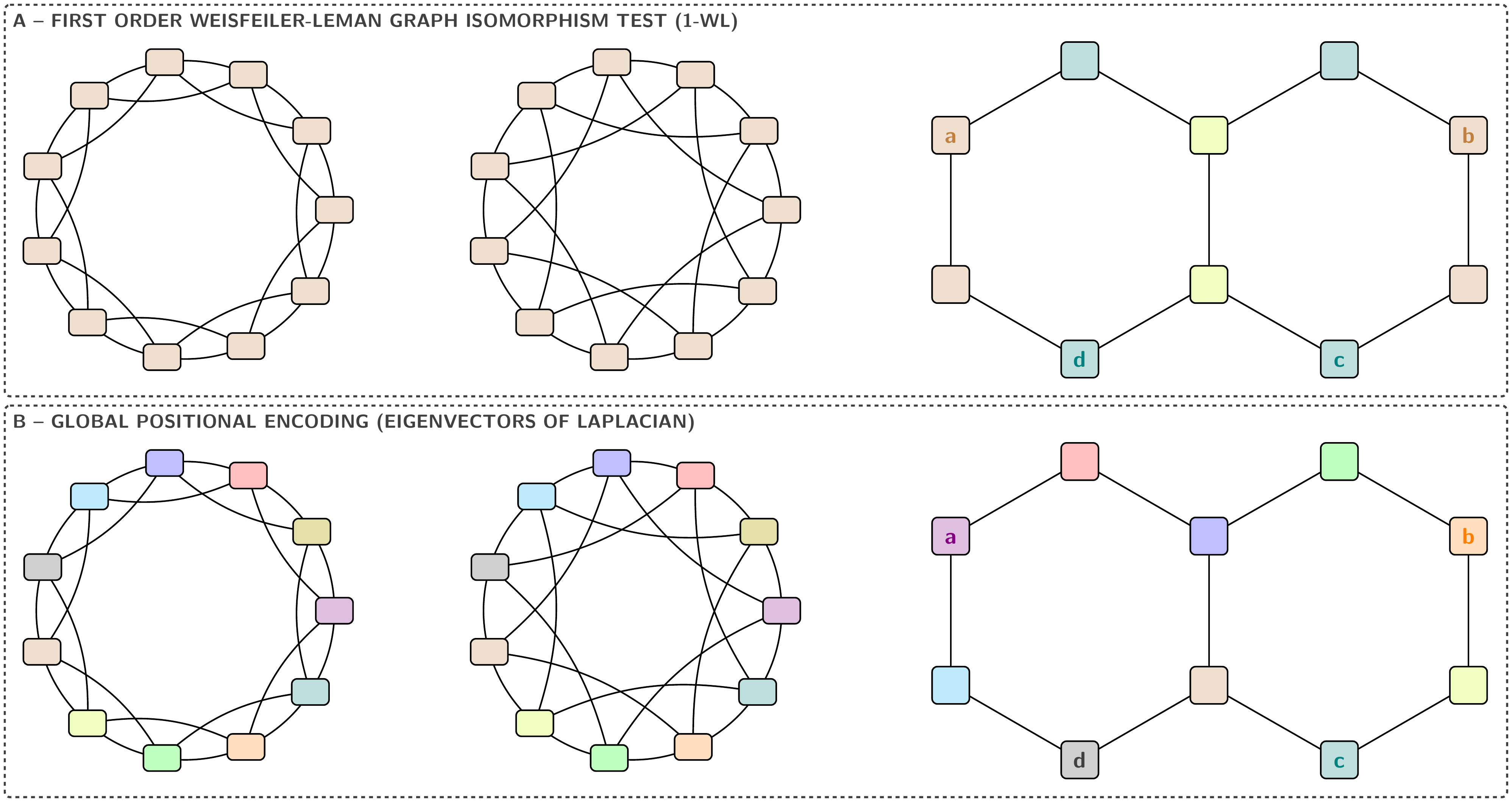}
            \caption{Showcasing the importance of global positional encoding for graph structured problems. \textit{(A)} In the left two columns the 1-WL incorrectly identifies the two graphs as being isomorphic. Similarly, it cannot distinguish for example node pairs (a,d) and (b,d) in the right column. \textit{(B)} When using a global positional encoding on the other hand these graphs and node pairs can be distinguished. Figure is a modified version of Figure~C.1 in~\citep{Rampasek:22}.}\label{fig:pe}
            \end{center}
        \end{figure}

        \section{G-Signatures can propagate global graph information in each step} \label{sec:globprop}
        \subsection{Global positional encoding}

        Prediction tasks on peptides molecular datasets as in \cite{Dwivedi:22} require a model to capture long-range interactions among atoms. For a given peptide they contain the peptide sequence, molecular graph, function, and 3D structure of the peptide. However, the graphs used for modeling the peptides correspond to 1D amino acid chains that do not have 2D or 3D peptide structure information included. This leads to large graphs with approx. 150 nodes each, and makes it important for the used ML model to identify the location of an amino acid in the graph. To enable this, a common candidate solution is global positional encoding \citep{Dwivedi:20} for each node. 

        To illustrate this more precisely we use the following simplified examples from \cite{Rampasek:22}. One of the tasks is to differentiate between two Circular Skip Link (CSL) graphs \citep{Murphy:19} as in the first two columns of \cref{fig:pe}. In the first row thereof it can be seen that the 1-Weisfeiler-Lehman test (1-WL) \citep{Weisfeiler:68} produces the same coloring of all nodes in the two non-isomorphic graphs. This means that the 1-WL cannot differentiate between them. This also holds for MPNNs since their expressive power is upper-bounded by 1-WL \citep{Xu:19,Morris:19}. This problem can be solved by using a global positioning encoding as shown in the second row. Similarly, in the third column for a Decalin molecular graph which has two rings of all Carbon atoms 1-WL (and thus MPNNs) would generate the same color for nodes $a, b$, and $c, d$, respectively. Thus, for link prediction the potential links $( a, d )$ and $(b, d)$ would be indistinguishable. Again, this issue can be solved by global positional encoding.

        However, for large graphs with many nodes MPNNs miss to propagate such global structural information between distant nodes. This is caused by the well-known over-squashing 
        phenomenon which is due to the local information aggregation mechanism for node updates.
        This means that crucial location information cannot be propagated by MPNNs. On the other hand, G-Signatures have access to this information as they do not rely on message passing and process the information for a given feature (and thus positional encoding) of all nodes at once.

        \subsection{Neighbors Match Problem}
        \begin{figure}[ht]
            \begin{center}
            \includegraphics[width=0.75\textwidth]{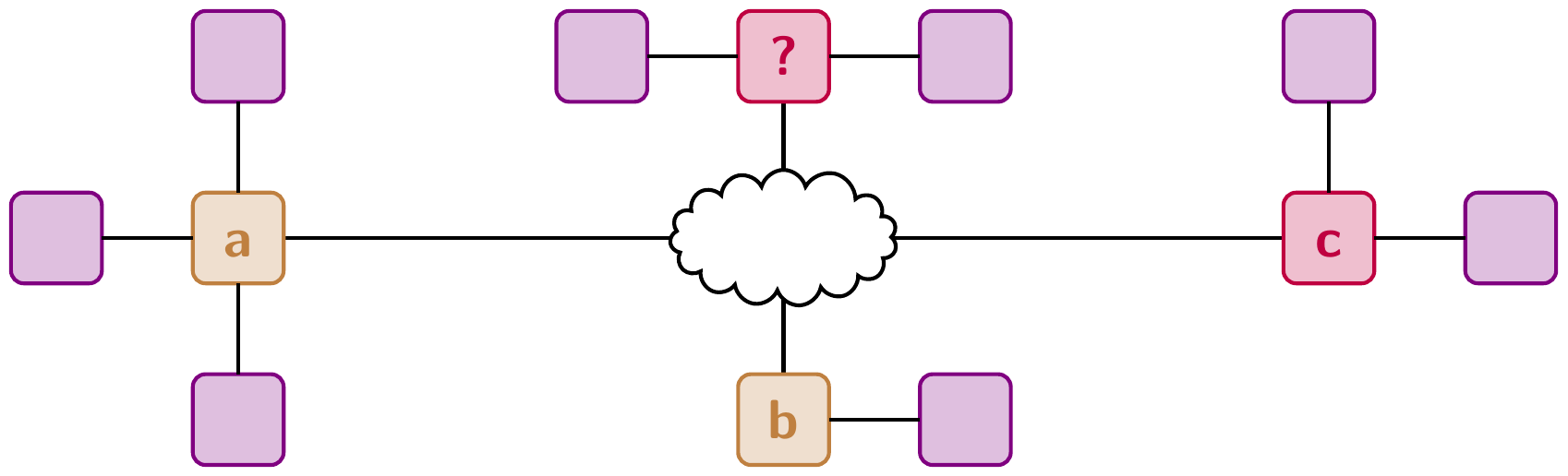}
            \caption{Showcasing the importance of global information propagation at the example of the neighbors match problem. To solve the task neighborhood information of distant nodes has to be propagated to the node marked with a question mark which can easily lead to over-squashing. Figure is a modified version of Figure~2 in~\citep{Alon:21}.}\label{fig:nmatch}
            \end{center}
        \end{figure}
        Another task where global graph propagation plays an important role is for real-world problems where computer program code is modeled as a graph \citep{Allamanis:18}. For example, long-range dependencies due to the usage of the same variable in distant locations are considered in this context. One such application would be the detection of variable misuses based for example on the variable types and values. 
        
        A simplified abstraction of this task is the Neighbors Match problem from \cite{Alon:21}. In the example illustrated in \cref{fig:nmatch} the goal is to predict the correct label of the target node marked with a question mark. In this case the correct label is "c" since the node with label "c" has the same number of neighbors as the node in question. For large graphs this task can easily lead to over-squashing in MPNNs, since the information from all nodes has to be propagated to the target node, whereas G-Signatures can immediately process this information for the entire graph.

        \section{Surrogate Models for Stochastic PDEs} \label{app:pde}
        
        \begin{figure}[ht]
            \begin{center}
            \includegraphics[width=0.782\textwidth]{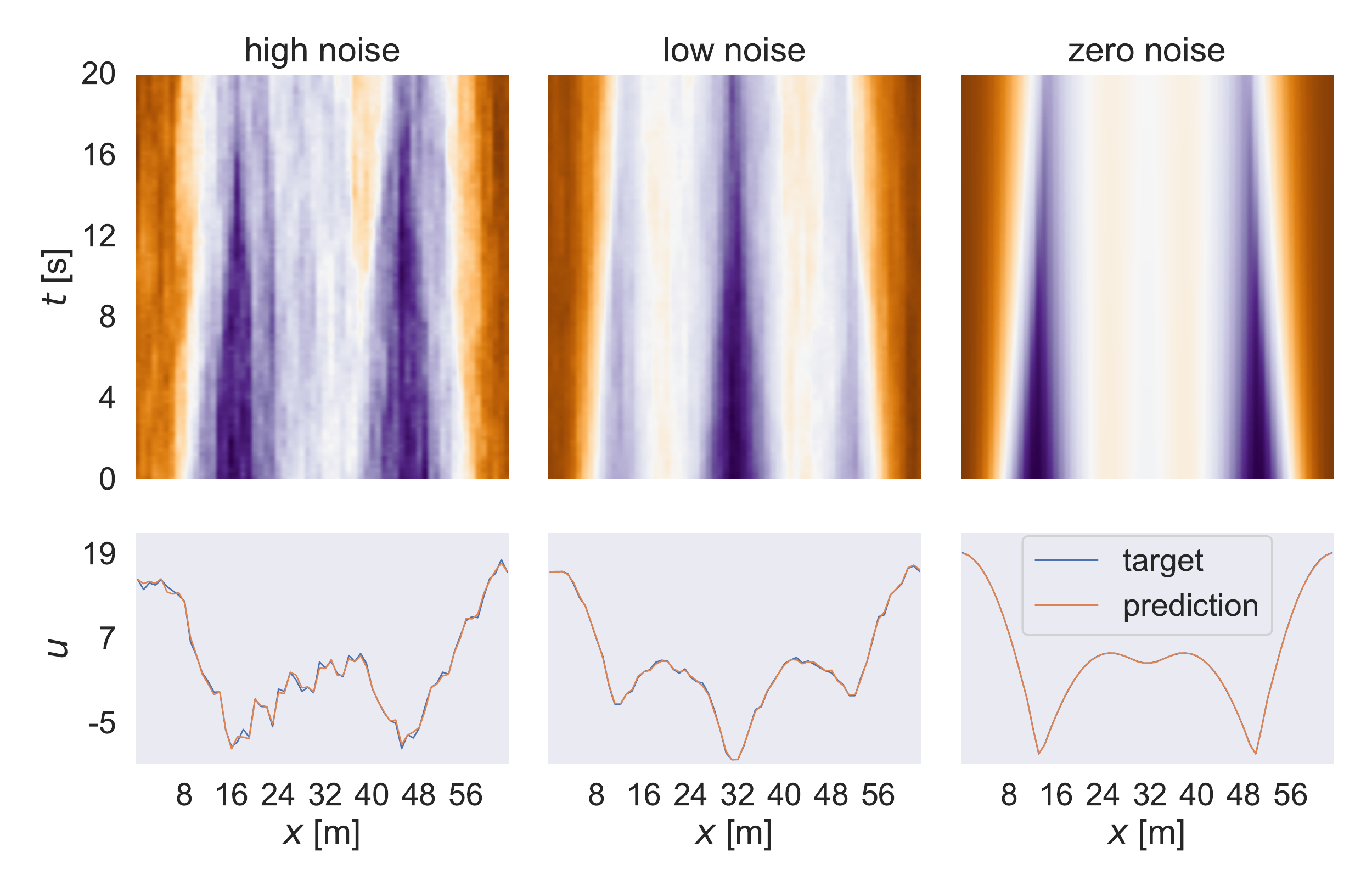}
            \caption{
            First row: Visualization of the dynamics of the KPZ equation unrolled over time. Second row: Exemplary comparison of G-Signatures predictions to target values at a fixed timestep of the above trajectory.}\label{fig:results_kpz}
            \end{center}
        \end{figure}
        
        The term $\frac{\lambda}{2} (\nabla \bm{u})^2$ in \cref{eq:kpz} is responsible for modeling shock waves.
        In one spatial dimension the KPZ equation corresponds to a stochastic version of the viscous Burgers' equation 
        -- a widely used PDE which occurs e.g., in fluid mechanics, or traffic modeling.
        Similarly, 
        larger values for $\nu$ have a smoothing effect, and both
        smaller values for $\nu$ and larger values for $\lambda$ result in larger shock formations. 
        The noise component $\eta(x,t)$ presents an additional difficulty
        in modeling the dynamics of \Cref{eq:kpz}.
        We aim to learn the mapping of $\bm{u}(x,t)$ to later points in time,
        We consider a modified implementation of the \textit{py-pde} package \citep{Zwicker:20}, obtaining data on $n_x =64$ spatial grid with temporal resolution of $\Delta t=0.2$. The data is split into $512$ training trajectories, $128$ validation trajectories, and $128$ test trajectories.
        Each time point of the respective training trajectories is used as individual starting point to obtain one training sample.
        The task is to predict the dynamics $10$ timesteps into the future given $10$ input timesteps, where performance is measured using the mean squared error (MSE).

        \section{Estimated Time of Arrival} \label{app:eta}
        We use the Floyd-Warshall \citep{Floyd:62,Warshall:62} algorithm of the Python package \textit{NetworkX}~\citep{Hagberg:08} to obtain ground truth values for the travel time along the shortest path between two nodes.
        Each node has 3 features and we use the edge embedding of \Cref{eq:embeddings} setting $m=3$ such that the network inputs for e.g. a graph with $500$ nodes are in $\dR^{3 \times 500}$, which is the input to G-Signatures.
        We split each dataset in a random way, such that \SI{512} graphs contribute to the training set, \SI{128} graphs to the validation set, and \SI{128} graphs to the test set.
        During training, each graph of the training set is augmented by applying a random permutation with respect to its nodes. This increases the potential total count of differently arranged training graphs by a factor of $\{500!, 1000!, 2000!\}$ for the respective ETA dataset.
        \newcommand{\SIETA}[2]{{\multirow{1}{*}{\SI[tight-spacing=false]{#1}}$\pm$ {\multirow{1}{*}{\SI[tight-spacing=false]{#2}}}}}
        \begin{table}[ht]
            \begin{center}
                \begin{tabular}{
                    S[table-format=4.0]@{\hskip 10pt}
                    S[table-format=1.2e1]@{\hskip 10pt}
                    c@{\hskip 10pt}
                    c@{\hskip 10pt}
                    c}
                \toprule
                {\makecell{\bf{}\#{}Nodes}} &
                {\makecell{\bf \#{}Edges}} &
                {\makecell{\bf GGCNs[MSE]}} &
                {\makecell{\bf GTs[MSE]}} &
                {\makecell{\bf G-Signatures[MSE]}}\\
                \midrule
                          & 2.50e+5 & \SIETA{1.936e-5}{0.242e-8} & \SIETA{1.951e-5}{9.466e-8} & \SIETA{1.979e-5}{0.523e-8} \\
                      500 & 1.25e+5 & \SIETA{4.444e-4}{0.027e-7} & \SIETA{4.444e-4}{0.237e-7} & \SIETA{0.785e-4}{0.421e-7} \\
                          & 2.50e+4 & \SIETA{1.846e-2}{0.171e-5} & \SIETA{1.846e-2}{0.177e-5} & \SIETA{0.196e-2}{0.067e-5} \\
                \cmidrule{1-5}
                          & 1.00e+6 & \SIETA{0.490e-5}{0.983e-8} & \SIETA{0.489e-5}{1.108e-8} & \SIETA{0.493e-5}{0.206e-8} \\
                     1000 & 5.00e+5 & \SIETA{1.317e-4}{0.304e-7} & \SIETA{1.317e-4}{0.168e-7} & \SIETA{0.197e-4}{0.024e-7} \\
                          & 1.00e+5 & \SIETA{0.548e-2}{0.069e-5} & \SIETA{0.548e-2}{0.001e-5} & \SIETA{0.049e-2}{0.008e-5} \\
                \cmidrule{1-5}
                          & 4.00e+6 & \SIETA{0.123e-5}{0.229e-8} & \SIETA{0.126e-5}{2.221e-8} & \SIETA{0.124e-5}{0.016e-8} \\
                     2000 & 2.00e+6 & \SIETA{0.383e-4}{0.014e-7} & \SIETA{0.383e-4}{0.094e-7} & \SIETA{0.049e-4}{0.010e-7} \\
                          & 4.00e+5 & \SIETA{0.163e-2}{0.018e-5} & \SIETA{0.163e-2}{0.003e-5} & \SIETA{0.012e-2}{0.001e-5} \\
                \bottomrule
                \end{tabular}
            \end{center}
            \caption{Results on the estimated time of arrival (ETA) datasets. The performance is reported by the mean squared error (MSE) with the standard error of the mean as the deviation. G-Signatures, Gated GCNs, and Graph Transformers are compared for different graph sizes, and different sparsity levels. When increasing the sparsity (reducing the number of edges) of the graphs, G-Signatures substantially outperform the competitors, often by an order of magnitude.}\label{tab:results_eta_mse}
        \end{table}

        \pagebreak
        \section{Latent Space Path Mapping} \label{app:lspm}
        \begin{figure}[h!]
            \begin{center}
            \includegraphics[width=0.7\textwidth]{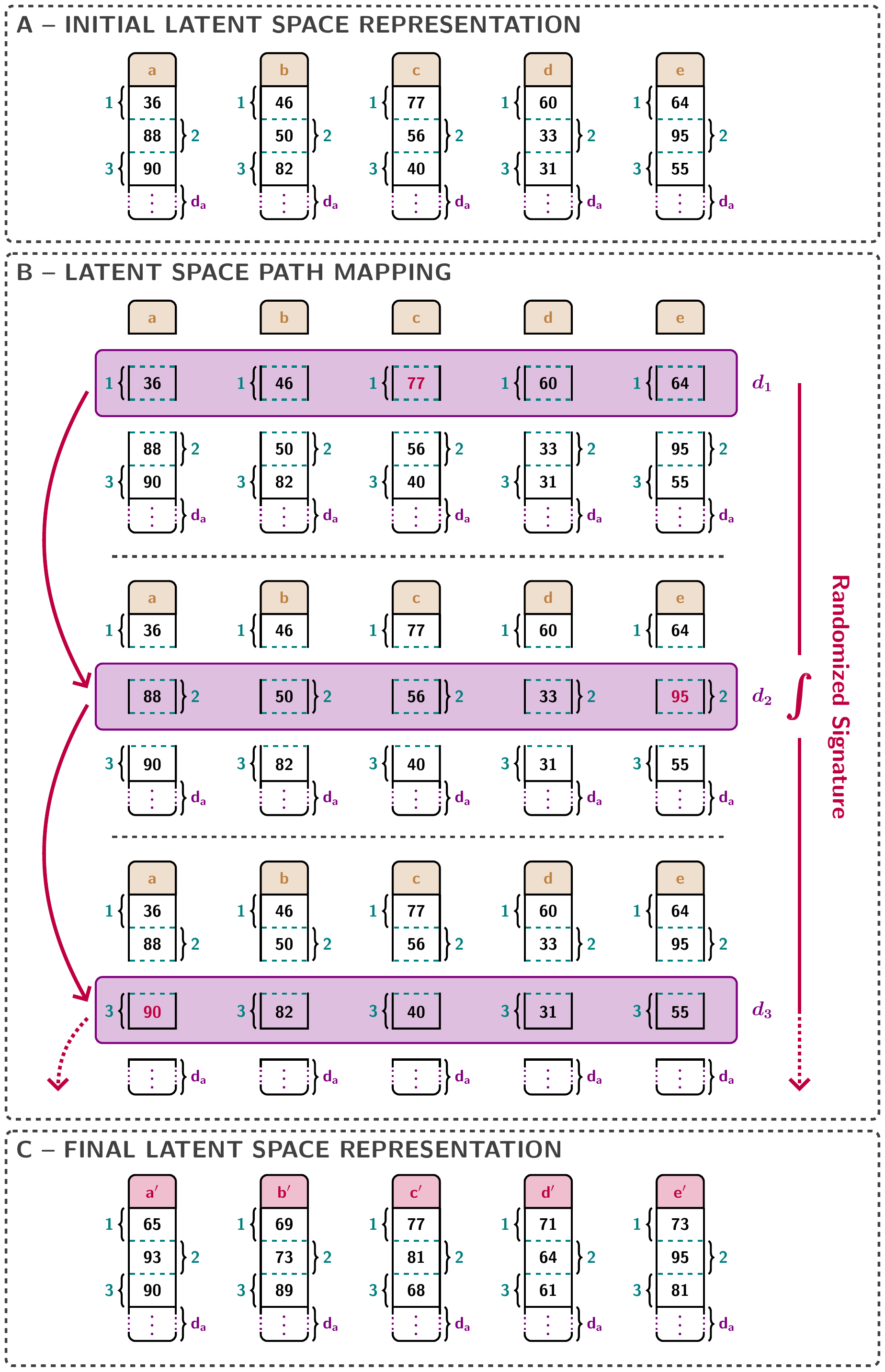}
            \caption{Latent space path mappings (LSPMs) traverse the initial latent space representation of a graph iteratively on a per-feature basis by applying randomized signature layers. In each iteration the randomized signature layers access global information of the entire latent graph in a parallel fashion, thus mitigating any negative smoothing and squashing effects. Exemplary, we showcase finding the maximum feature value of all latent nodes in purple. As our method operates on all latent nodes in parallel, the maximum value per feature is found in a single step.}\label{fig:lspm_illustrated}
            \end{center}
        \end{figure}
        
        \begin{figure}[ht]
            \begin{center}
            \includegraphics[width=\textwidth]{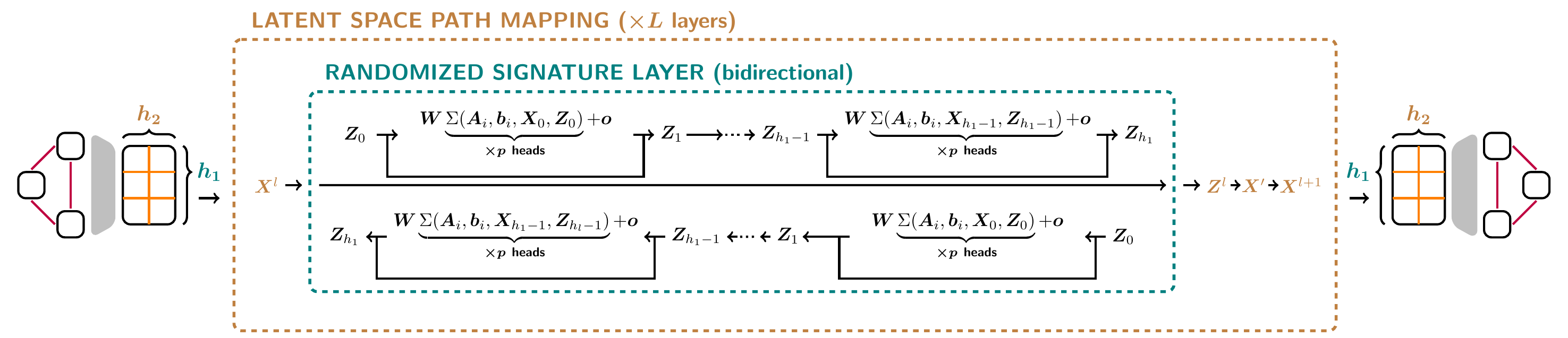}
            \caption{
            Illustration of the G-Signatures architecture. Graph conversion allows us to map graph structured data into a path $X:[0,T] \rightarrow \dR^{h_2}$ in the latent space sampled at $h_1$ points. This constitutes the latent space representation of a graph. Latent space path mappings (LSPM) process this latent representation sequentially via stacked randomized signature layers. The decoder maps the latent graph representation to a task dependent output space.}\label{fig:architecture_overview}
            \end{center}
        \end{figure}

        \newpage
        \appendix
        \onecolumn

        \setcounter{secnumdepth}{1}
        \setcounter{section}{6}
        \section{Ablation Studies} \label{app:ablat}
        We compare results of G-Signatures against ablated versions thereof on the KPZ dataset with a low noise level. We investigate the effect of all combinations of our adjustments.
        We apply the same hyperparameter search procedure as for the main experiments of G-Signatures. Each entry in \cref{tab:ablat_kpz} shows the corresponding result by omitting a certain combination of adjustments. The diagonal elements of both sub-tables indicate the omission of the respective row/column element, e.g., a combination of {\sc Sparsity} and {\sc Sparsity} corresponds to the omission of just the sparsity adjustment, a combination of {\sc Sparsity} and {\sc Initialization} corresponds to the omission of the sparsity, as well as the initialization adjustments. The "$\cdots$" serves as a placeholder since the lower sub-table is symmetric, and "\----{}" indicates that the given combination is not applicable. The {\sc Trainability} ablation refers to a fixed randomized signature, i.e. $A_i$ and $b_i$ are kept fixed. It can be clearly seen that each adjustment is necessary to improve the performance of G-Signatures, and that omitting them steadily worsens the results for all of their combinations.
                
        \newcommand{\SIEABL}[2]{{\multirow{1}{*}{\SI[tight-spacing=false]{#1}}$\pm$ {\multirow{1}{*}{\SI[tight-spacing=false]{#2}}}}}
        \begin{table}[ht]
            \begin{center}
                \begin{tabular}{
                    c
                    c
                    c
                    c}
                \toprule
                    {\makecell{\bf Ablations}}      &
                    {\makecell{\sc None (ours)}}    &
                    {\makecell{\sc Trainability}}    &
                    {\makecell{\sc All (original)}} \\
                \midrule
                    None (ours)      & \SIEABL{0.980e-3}{0.007e-3} & \----{}                     & \----{}  \\
                    Trainability     & \----{}                     & \SIEABL{1.022e-3}{0.005e-3} & \----{} \\
                    All (original)   & \----{}                     & \----{}                     & \SIEABL{1.308e-3}{0.018e-3} \\
                \midrule
                    {\makecell{\bf Ablations}}      &
                    {\makecell{\sc Sparsity}}       &
                    {\makecell{\sc Initialization}} &
                    {\makecell{\sc Activation}}     \\
                \midrule
                    Sparsity         & \SIEABL{0.983e-3}{0.001e-3} & $\cdots$ & $\cdots$ \\
                    Initialization   & \SIEABL{0.994e-3}{0.008e-3} & \SIEABL{0.987e-3}{0.007e-3} & $\cdots$ \\
                    Activation       & \SIEABL{0.996e-3}{0.009e-3} & \SIEABL{1.003e-3}{0.005e-3} & \SIEABL{0.990e-3}{0.005e-3} \\
                \midrule
                    Sigmoid          & \----{}                     & \----{}                     & \SIEABL{0.993e-3}{0.004e-3} \\
                    TanH             & \----{}                     & \----{}                     & \SIEABL{0.998e-3}{0.004e-3} \\
                \bottomrule
                \end{tabular}
            \end{center}
            \caption{
            Performance results of G-Signatures compared to its ablated versions on the KPZ dataset with a low noise level. The reported deviation is the standard error of the mean. G-Signatures perform best with all LSPM adjustments applied.
            }\label{tab:ablat_kpz}
        \end{table}

\begin{figure}[ht]
            \begin{center}
            \includegraphics[width=\linewidth]{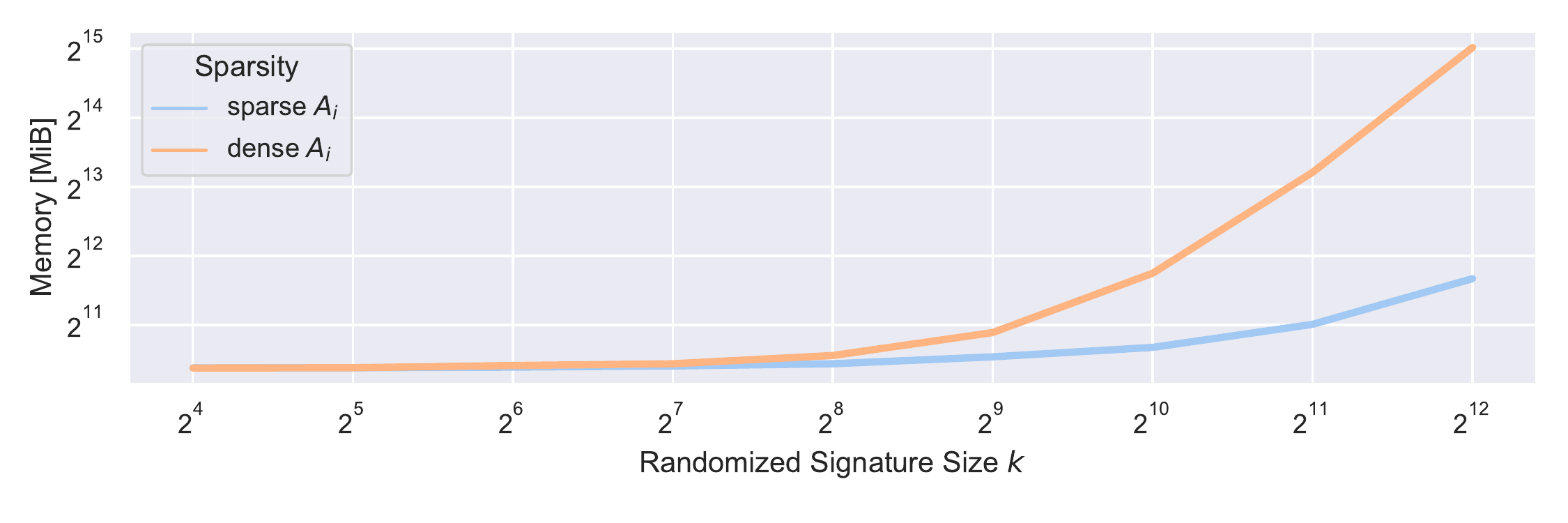}
            \caption{Memory consumption in MiB of G-Signatures on an NVIDIA A100 GPU 40GB for a single sample during training. All hyperparameters are kept fixed, except size $k$ of $A_i$ and $b_i$ of the randomized signature layer (see~\cref{alg:randomized} for details).}\label{fig:memory_consumption}
            \end{center}
        \end{figure}

        It is notable that when omitting the sparsity adjustment, i.e., using dense matrices for $A_{i}$, both the performance decreases and the method is much less memory- and compute-efficient. More precisely, there is a quadratic instead of a linear scaling for $A_{i}$, as can be seen in~\cref{fig:memory_consumption}. 
        
\begin{figure}[ht]
            \begin{center}
            \includegraphics[width=\linewidth]{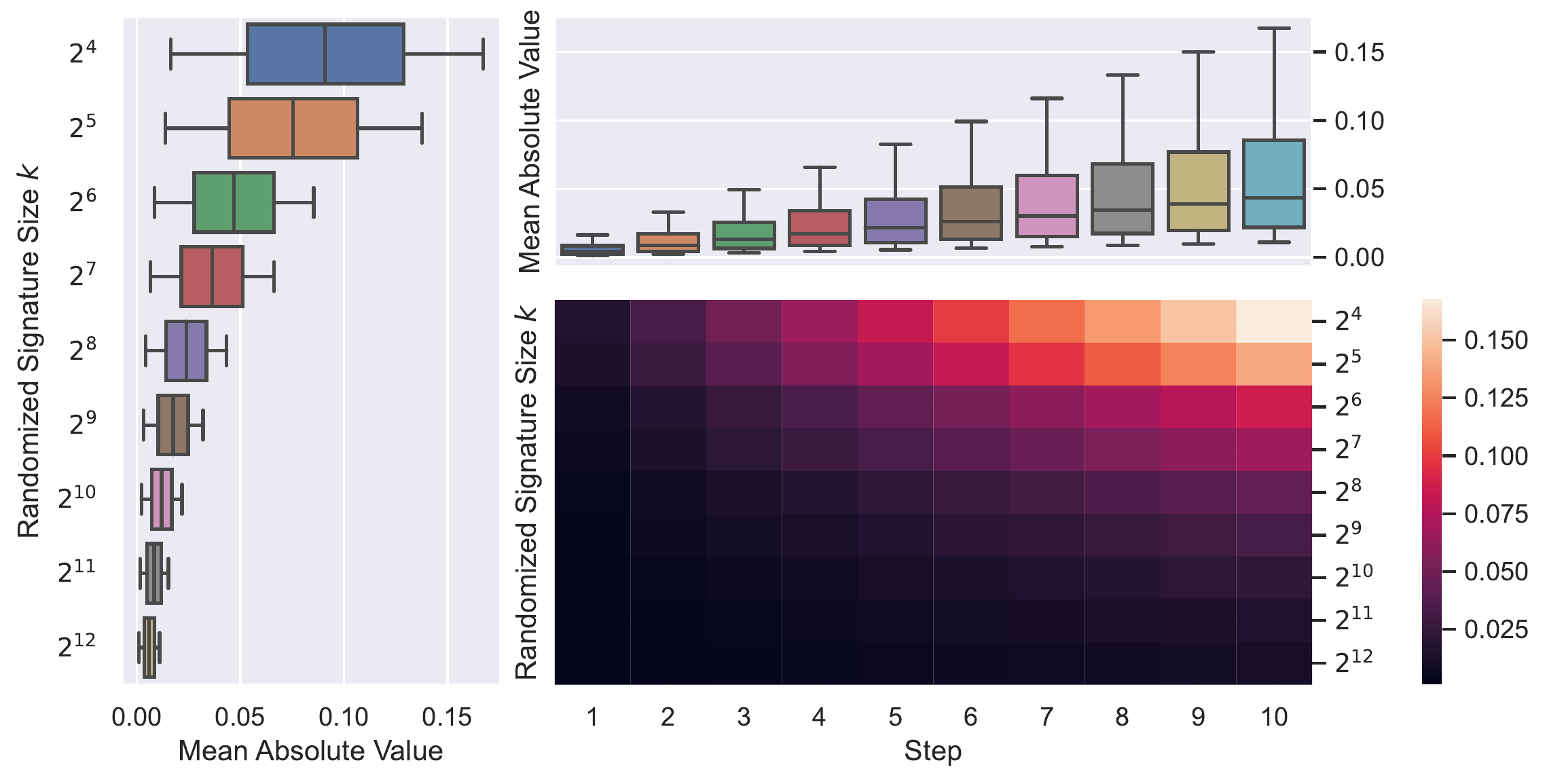}
            \caption{Ablation {\sc None (ours)}. Lower right: Mean Absolute Values (MAVs) of randomized signature computation per step and randomized signature size. Upper right: Range of MAVs across randomized signature sizes per step. Left: Range of MAVs across steps per randomized signature size. Our modified version does not suffer from exploding signals.}\label{fig:analysis_none}
            \end{center}
        \end{figure}

\begin{figure}[ht]
            \begin{center}
            \includegraphics[width=\linewidth]{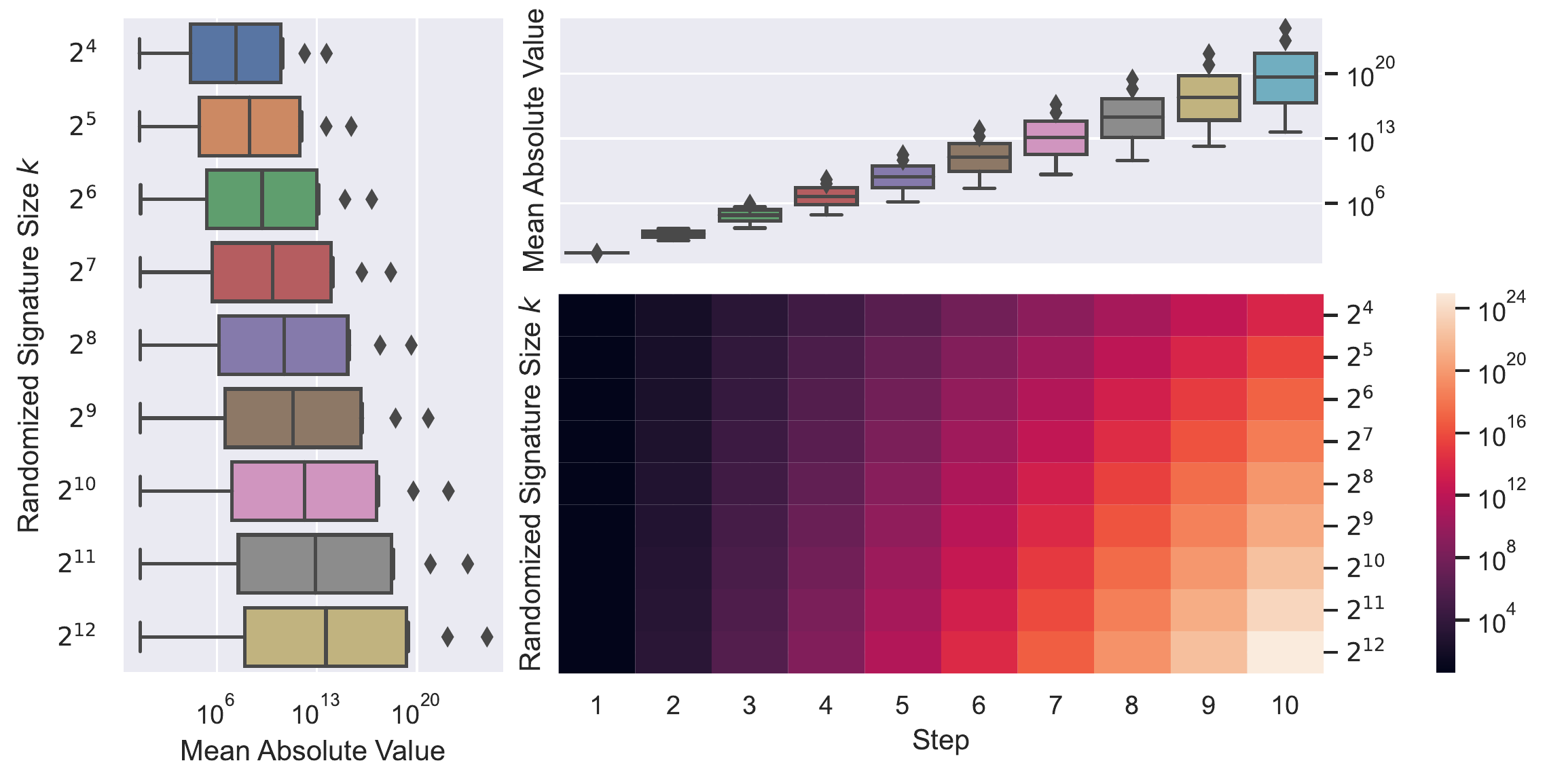}
            \caption{Ablation {\sc All (original)}. Lower right: Mean Absolute Values (MAVs) of randomized signature computation per step and randomized signature size. Upper right: Range of MAVs across randomized signature sizes per step. Left: Range of MAVs across steps per randomized signature size. The original computation of the randomized signature immediately produces exploding values.}\label{fig:analysis_all}
            \end{center}
        \end{figure}

\begin{figure}[ht]
            \begin{center}
            \includegraphics[width=\linewidth]{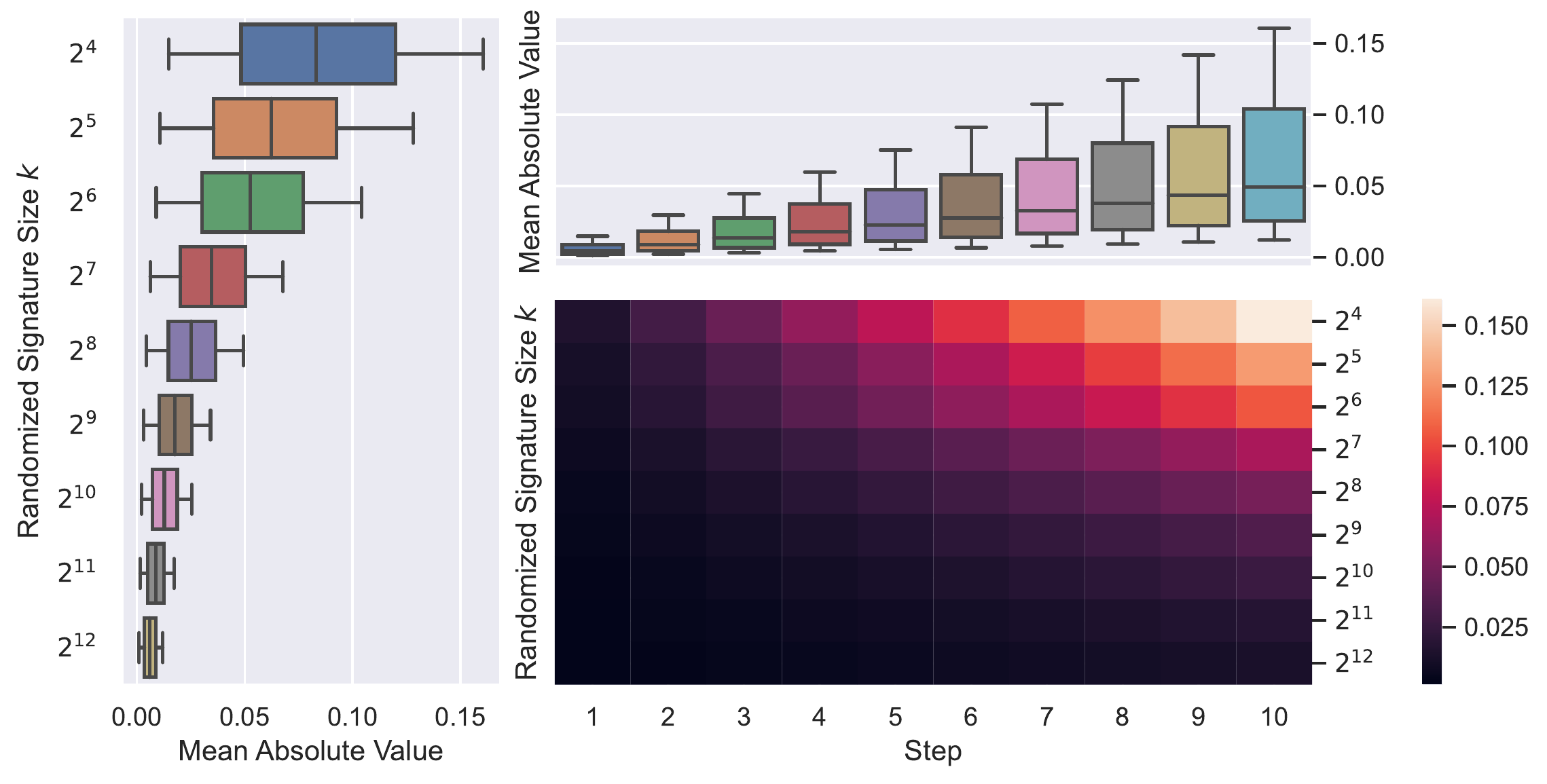}
            \caption{Ablation {\sc Sparsity}. Lower right: Mean Absolute Values (MAVs) of randomized signature computation per step and randomized signature size. Upper right: Range of MAVs across randomized signature sizes per step. Left: Range of MAVs across steps per randomized signature size. Omitting the sparsity adjustment keeps computation stable at the expense of memory.}\label{fig:analysis_sparsity}
            \end{center}
        \end{figure}

\begin{figure}[ht]
            \begin{center}
            \includegraphics[width=\linewidth]{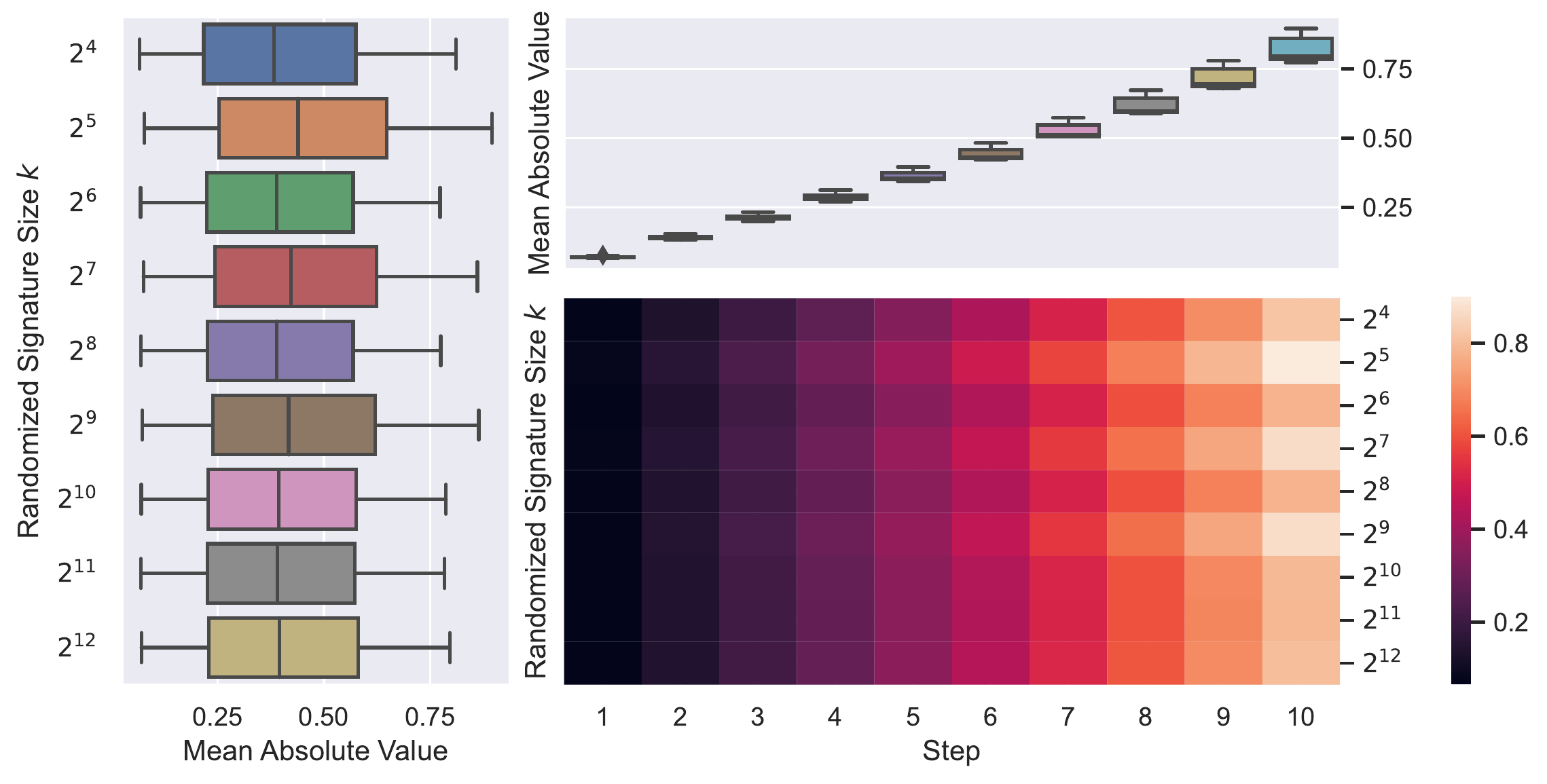}
            \caption{Ablation {\sc Initialization}. Lower right: Mean Absolute Values (MAVs) of randomized signature computation per step and randomized signature size. Upper right: Range of MAVs across randomized signature sizes per step. Left: Range of MAVs across steps per randomized signature size. Omitting the initialization adjustment results in a larger value magnitude.}\label{fig:analysis_initialization}
            \end{center}
        \end{figure}

\begin{figure}[ht]
            \begin{center}
            \includegraphics[width=\linewidth]{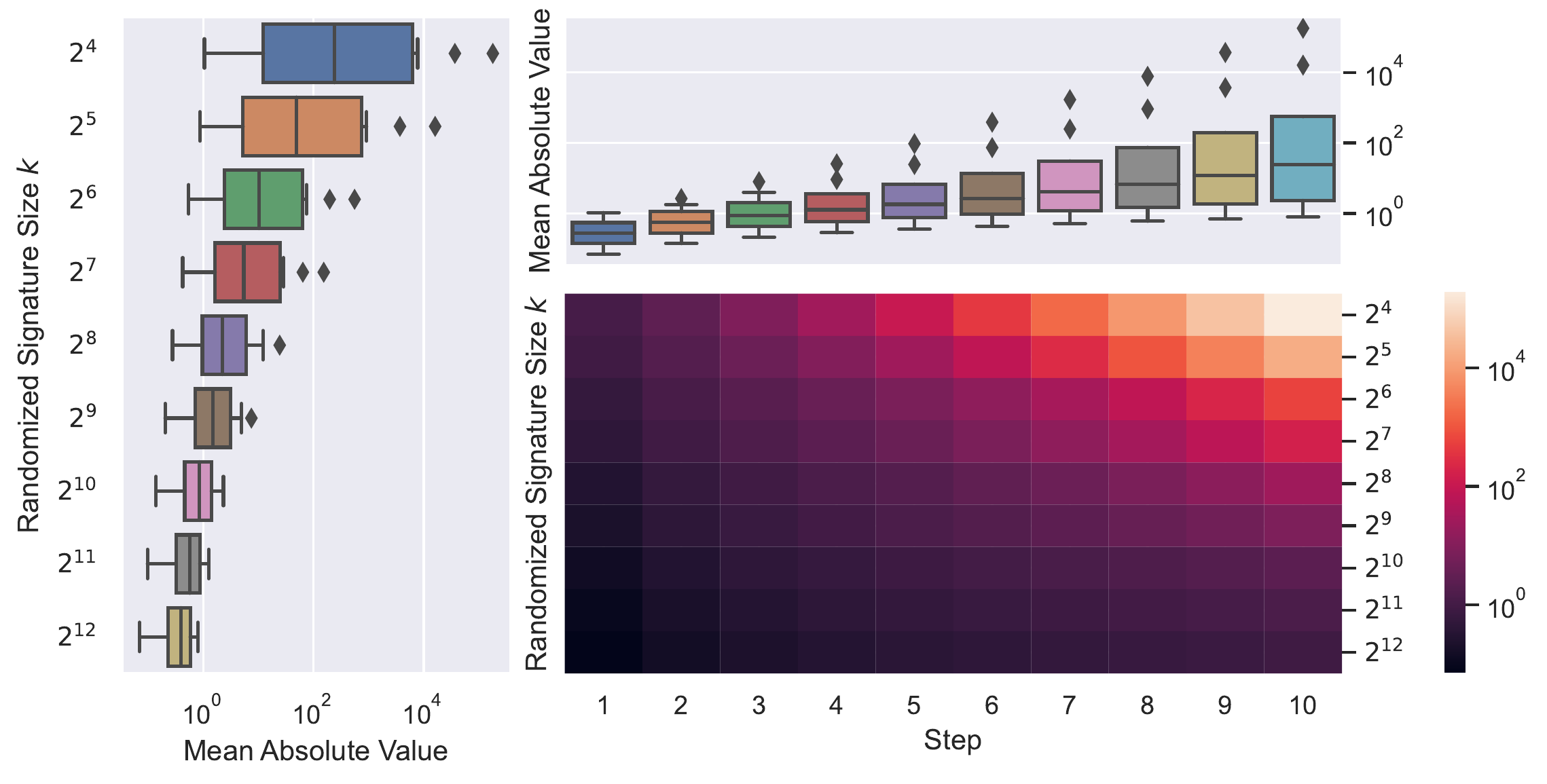}
            \caption{Ablation {\sc Activation}. Lower right: Mean Absolute Values (MAVs) of randomized signature computation per step and randomized signature size. Upper right: Range of MAVs across randomized signature sizes per step. Left: Range of MAVs across steps per randomized signature size. Omitting the activation adjustment renders the randomized signature computation susceptible to large signals.}\label{fig:analysis_activation}
            \end{center}
        \end{figure}

\begin{figure}[ht]
            \begin{center}
            \includegraphics[width=\linewidth]{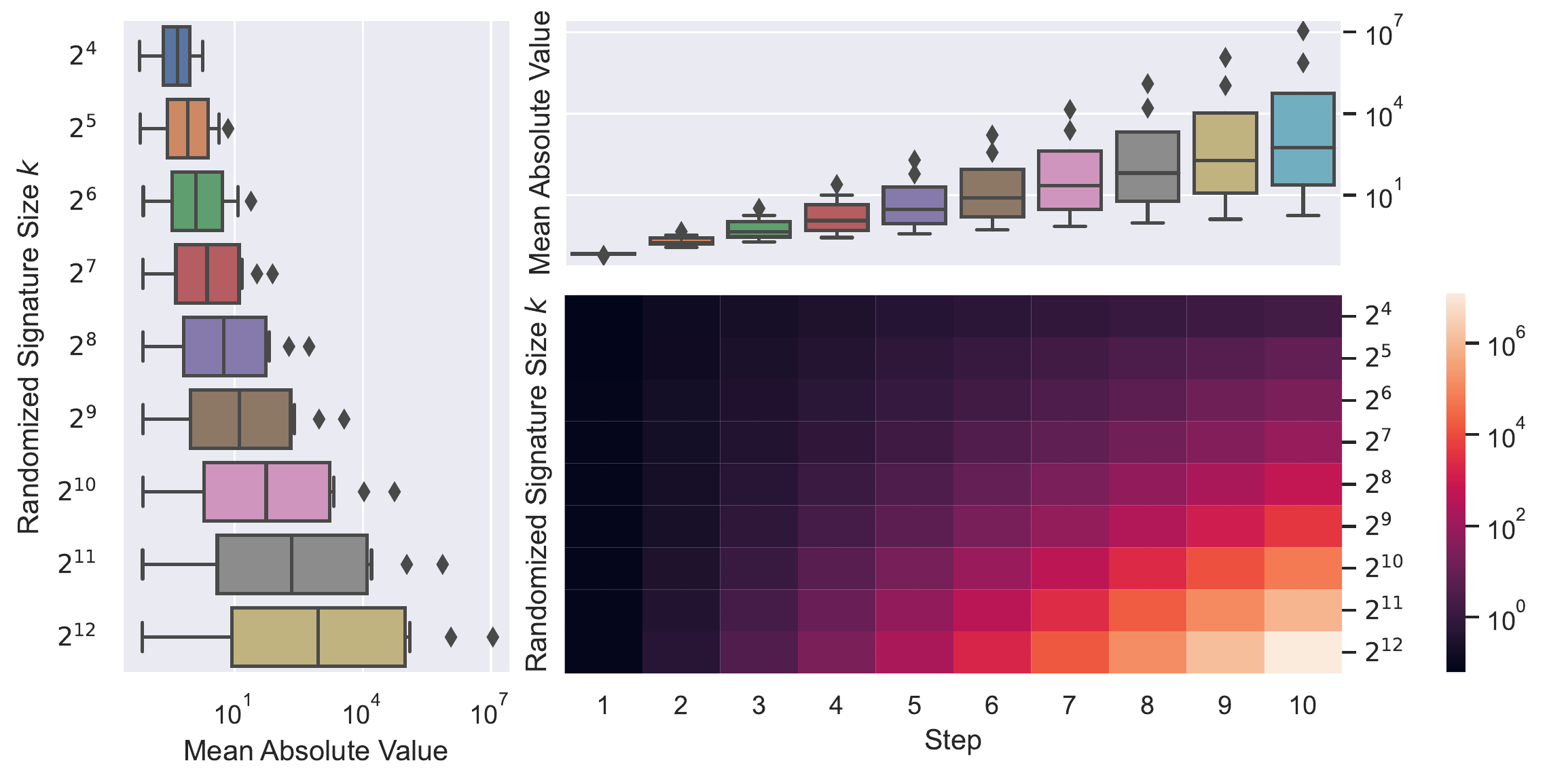}
            \caption{Ablation {\sc Initialization/Sparsity}. Lower right: Mean Absolute Values (MAVs) of randomized signature computation per step and randomized signature size. Upper right: Range of MAVs across randomized signature sizes per step. Left: Range of MAVs across steps per randomized signature size. Omitting both initialization and sparsity adjustments results in exploding signals.}\label{fig:analysis_initialization_sparsity}
            \end{center}
        \end{figure}

\begin{figure}[ht]
            \begin{center}
            \includegraphics[width=\linewidth]{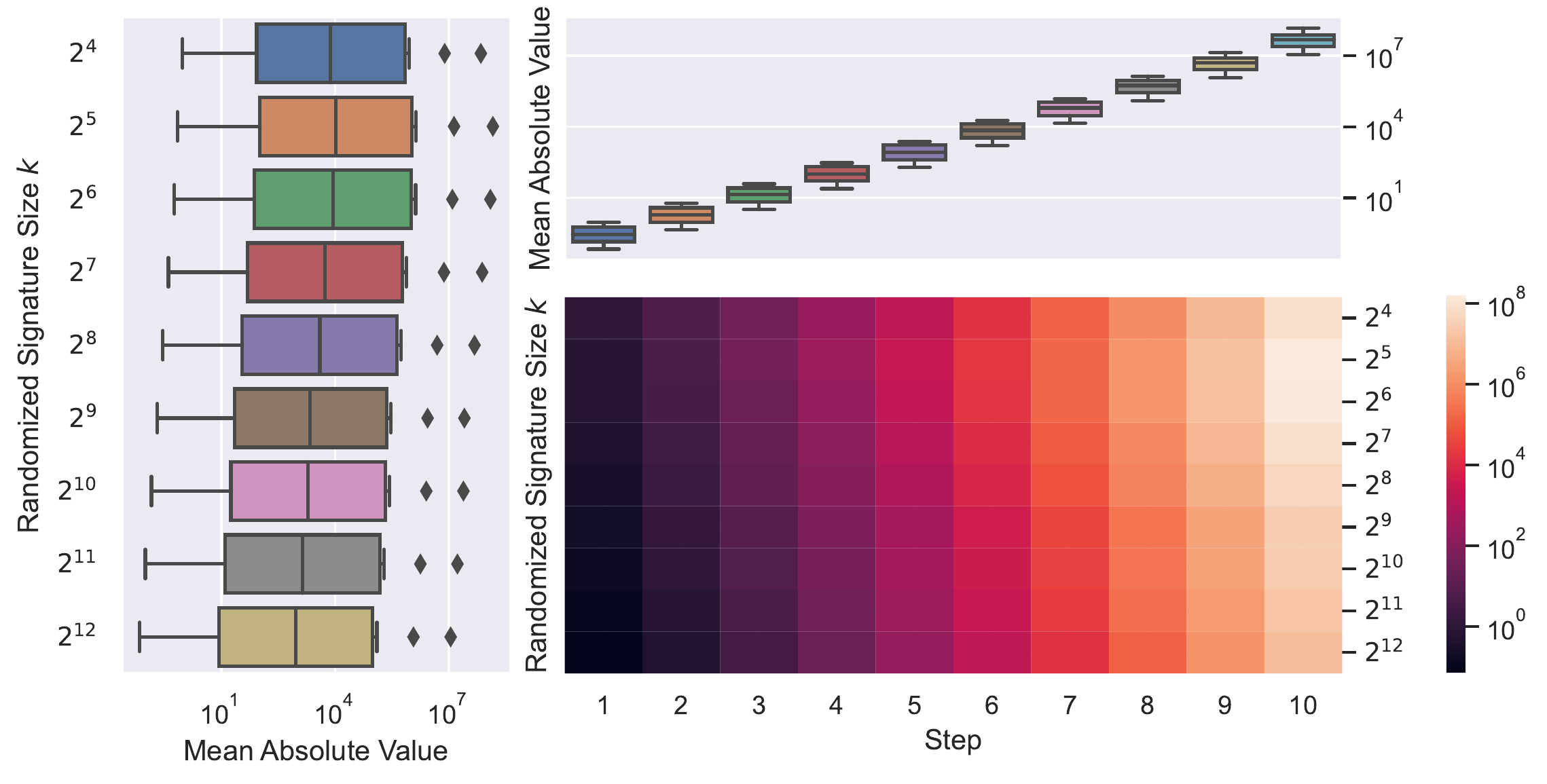}
            \caption{Ablation {\sc Activation/Sparsity}. Lower right: Mean Absolute Values (MAVs) of randomized signature computation per step and randomized signature size. Upper right: Range of MAVs across randomized signature sizes per step. Left: Range of MAVs across steps per randomized signature size. Omitting both activation and sparsity adjustments leads to exploding values.}\label{fig:analysis_activation_sparsity}
            \end{center}
        \end{figure}

\begin{figure}[ht]
            \begin{center}
            \includegraphics[width=\linewidth]{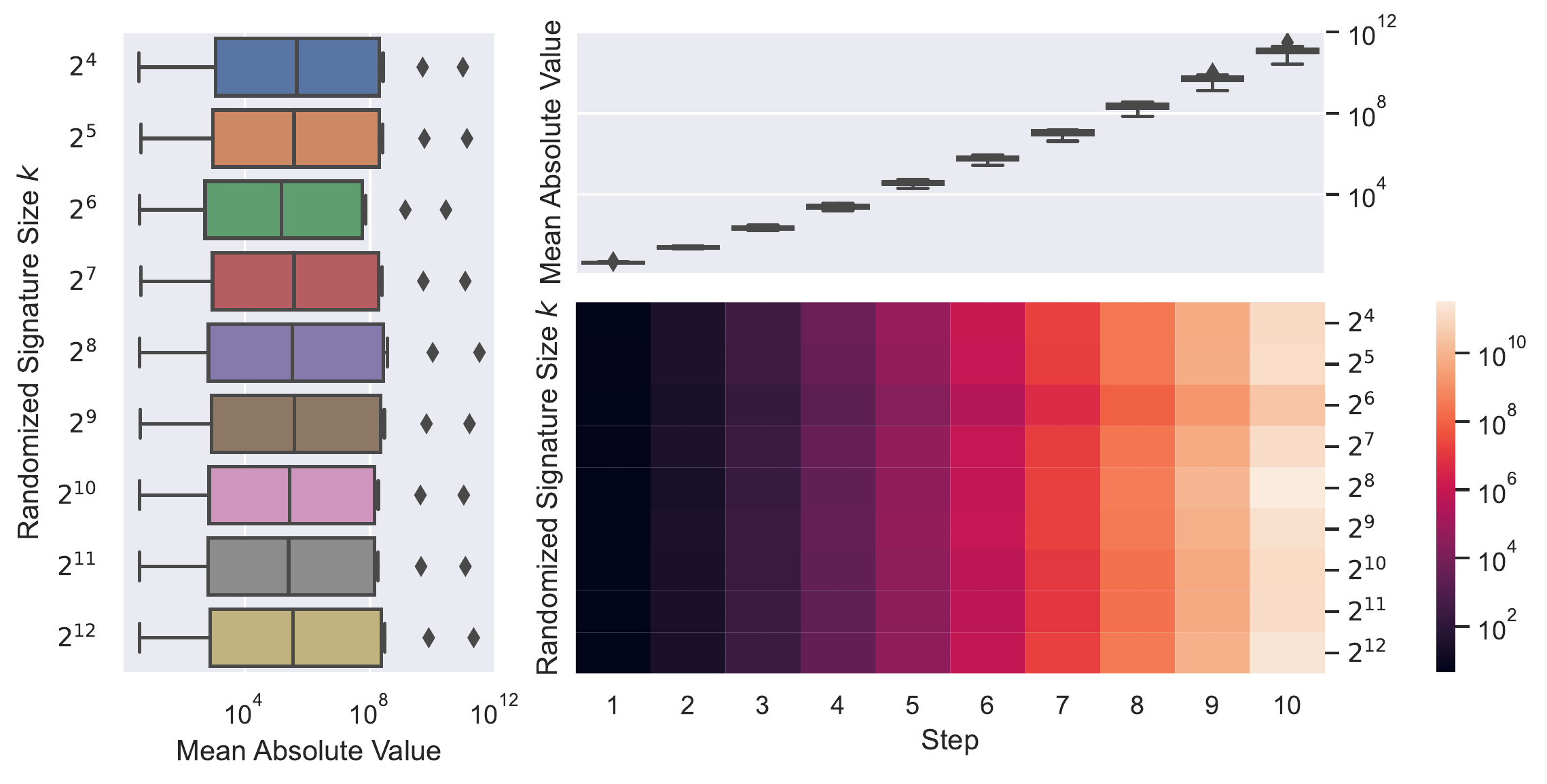}
            \caption{Ablation {\sc Initialization/Activation}. Lower right: Mean Absolute Values (MAVs) of randomized signature computation per step and randomized signature size. Upper right: Range of MAVs across randomized signature sizes per step. Left: Range of MAVs across steps per randomized signature size. Omitting both initialization and activation adjustments results in exploding signals.}\label{fig:analysis_initialization_activation}
            \end{center}
        \end{figure}

        Additionally, many hyperparameter settings of the ablated versions cannot even be trained successfully. We illustrate this in more detail starting from \cref{fig:analysis_none} to \cref{fig:analysis_initialization_activation}. For this, we investigate for all ablations in \cref{tab:ablat_kpz} the magnitude of the values computed by the signature based on \cref{alg:randomized}. In each of the figures the absolute values per timestep and signature size (averaged over samples) are plotted in the lower-right sub-figure. The range of these values across the signature sizes per step is visualized in the upper-right plot. The range across the steps per signature size is shown in the left sub-figure. It can be seen that in most cases the computation leads to exploding values after a few steps already.

        \begin{figure}[ht]
            \begin{center}
            \includegraphics[width=\linewidth]{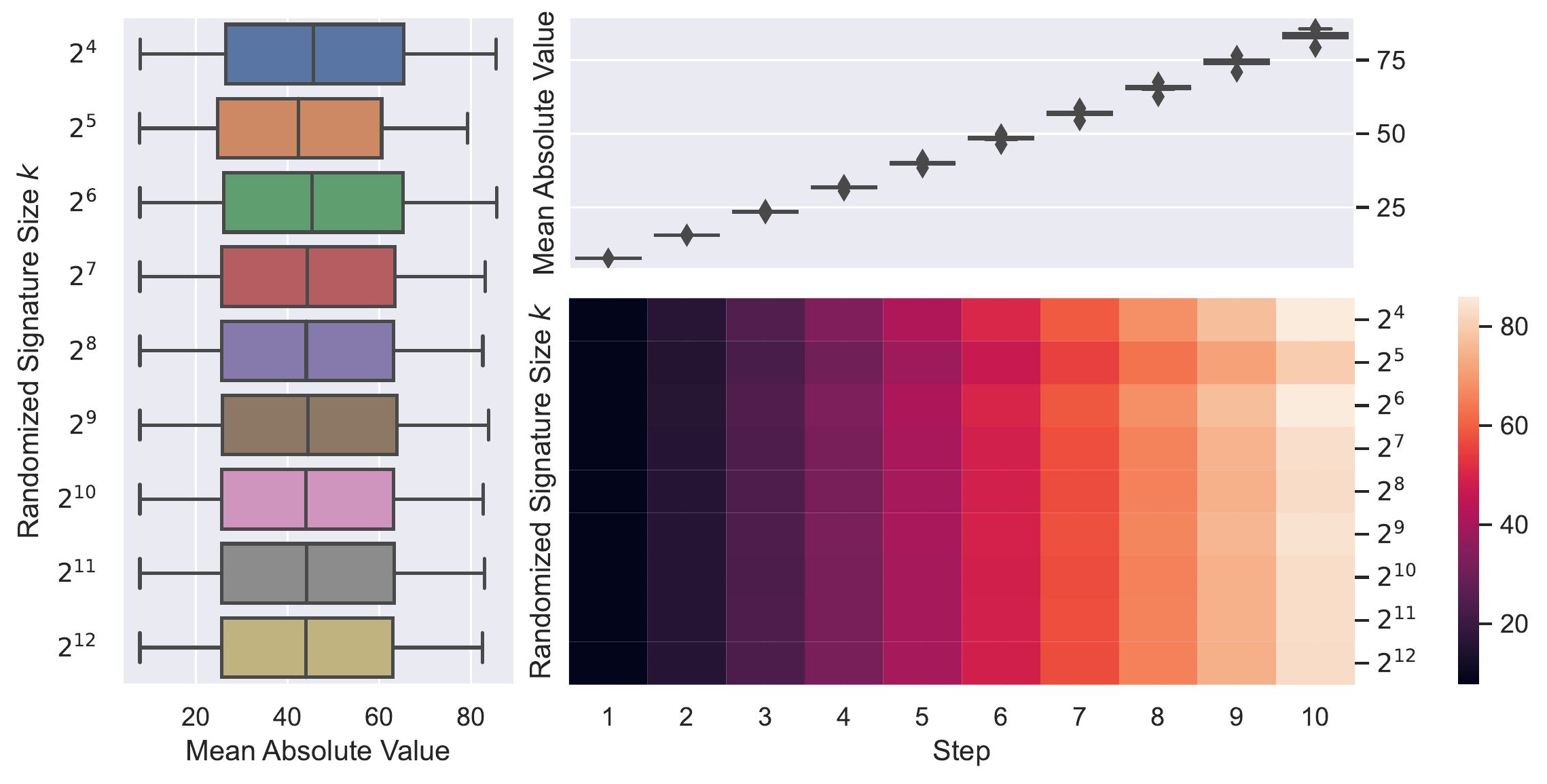}
            \caption{Ablation {\sc Activation (Sigmoid)}. Lower right: Mean Absolute Values (MAVs) of randomized signature computation per step and randomized signature size. Upper right: Range of MAVs across randomized signature sizes per step. Left: Range of MAVs across steps per randomized signature size. Exchanging our activation adjustment with a $sigmoid$ non-linearity renders the randomized signature computation susceptible to large signals. Additionally, the consistency of the MAVs indicate a string saturating effect.}\label{fig:analysis_activation_sigmoid}
            \end{center}
        \end{figure}

        \begin{figure}[ht]
            \begin{center}
            \includegraphics[width=\linewidth]{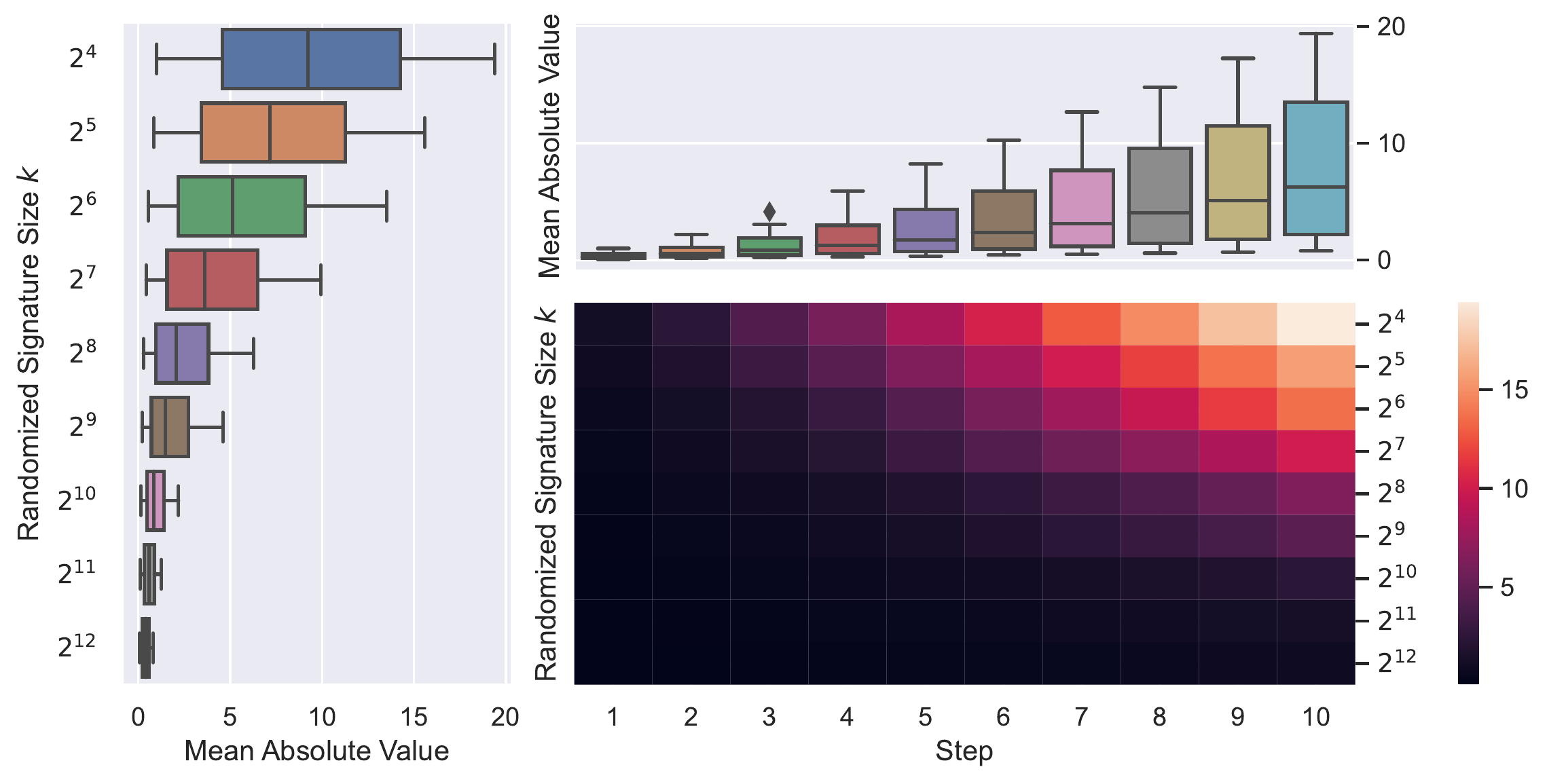}
            \caption{Ablation {\sc Activation (TanH)}. Lower right: Mean Absolute Values (MAVs) of randomized signature computation per step and randomized signature size. Upper right: Range of MAVs across randomized signature sizes per step. Left: Range of MAVs across steps per randomized signature size. Exchanging our activation adjustment with a $tanh$ non-linearity renders the randomized signature computation susceptible to large signals for small $k$. This effect is pronounced for higher step counts.}\label{fig:analysis_activation_tanh}
            \end{center}
        \end{figure}

        \newpage
        \appendix
        \onecolumn

        \setcounter{secnumdepth}{1}
        \setcounter{section}{7}
        \section{Hyperparameter Selection} \label{app:hyperparameters}
        For Gated GCNs and Graph Transformers we use hyperparameters that were already successfully applied to learn dependencies across longer node paths in \citep{Dwivedi:21,Dwivedi:22}. We complement this by a manual hyperparameter search which results in the following hyperparameter search spaces.
        
        \begin{table}[ht]
            \begin{center}
                \begin{tabular}{l c c c c}
                \toprule
              {\makecell{\bf {Method}}} & {\makecell{\bf \#{}Layers}} & {\makecell{\bf Hidden Size}} & {\makecell{\bf Learning Rate}} & {\makecell{\bf \#{}Heads}} \\
                \midrule
                    GGCNs & $\left\{1, \{2 + 2 i \}_{i=0}^5,16\right\}$ & $\left\{6,12,32,70,100,138\right\}$ & $\left\{10^{-\left\{2,3,4\right\}}, 5 \times 10^{-4}\right\}$ &  - \\
                    GTs & $\left\{1,2,4,8,10\right\}$ & $\left\{6,12,32,80,120\right\}$ & $\left\{10^{-\left\{2,3,4\right\}}, 5 \times 10^{-4}\right\}$ & $\left\{2^i\right\}_{i=0}^4$ \\
                    G-Sigs. & $\left\{1, 2, 3\right\}$ & $\left\{32 + i\right\}_{i=0}^{64}$ & $10^{\left[-3,-1\right]}$ & $\left\{1,2,3\right\}$ \\
                \bottomrule
                \end{tabular}
            \end{center}
            \caption{Hyperparameter settings for compared methods on CoMA.}
        \end{table}
        \begin{table}[ht]
            \begin{center}
                \begin{tabular}{l c c c c}
                \toprule
              {\makecell{\bf {Method}}} & {\makecell{\bf \#{}Layers}} & {\makecell{\bf Hidden Size}} & {\makecell{\bf Learning Rate}} & {\makecell{\bf \#{}Heads}}\\
                \midrule
                    GGCNs & $\left\{1,3,4,6,8,10,12\right\}$ & $\left\{6,12,32,70,100\right\}$ & $\left\{10^{-\left\{2,3,4\right\}}, 5 \times 10^{-4}\right\}$ & -\\
                    GTs & $\left\{1,3,4,6,8,10\right\}$ & $\left\{6,12,32,80\right\}$ & $\left\{10^{-\left\{2,3,4\right\}}, 5 \times 10^{-4}\right\}$ & $\left\{1,2,4,8\right\}$\\
                    G-Sigs. & $\left\{1, 2, 3\right\}$ & $\left\{32 + i\right\}_{i=0}^{64}$ & $10^{\left[-3,-1\right]}$ & $\left\{1,2,3\right\}$\\
                \bottomrule
                \end{tabular}
            \end{center}
            \caption{Hyperparameter settings for compared methods on ETA datasets.}
        \end{table}        
        \begin{table}[ht]
            \begin{center}
                \begin{tabular}{l c c c c c}
                \toprule
              {\makecell{\bf {Method}}} & {\makecell{\bf \#{}Layers}} & {\makecell{\bf Hidden Size}} & {\makecell{\bf Learning Rate}} & {\makecell{\bf \#{}Heads}} & {\makecell{\bf \#{}Modes}}\\
                \midrule
                    ResNets & $\left\{4,8,16\right\}$ & $\left\{32,64,128\right\}$ & $10^{\left[-3,-1\right] }$ & - & -\\
                    FNOs & $\left\{3,4,5\right\}$ & $\left\{32,128,256\right\}$ & $10^{\left[-3,-1\right] }$ & - & $\left\{8,12,16,32\right\}$\\
                    G-Sigs. & $\left\{1, 2, 3\right\}$ & $\left\{32 + i\right\}_{i=0}^{64}$ & $10^{\left[-3,-1\right]}$ & $\left\{1,2,3\right\}$ & -\\
                \bottomrule
                \end{tabular}
            \end{center}
            \caption{Hyperparameter settings for compared methods on KPZ datasets.}
        \end{table}
        All methods are trained for 1000 epochs with early stopping.
        The GNN methods are trained with either batch~\citep{Szegedy:15}, or layer normalization~\citep{Ba:16},  and by setting Dropout~\citep{Hinton:12} to $0.0$ or $0.1$. We test both averaging and summing over all nodes for the readout function of the respective GNNs. For all methods we use the LAMB \citep{You:20} optimizer with an optional cosine annealed learning rate scheduling \citep{Loshilov:16}. The signature size hyperparameter of G-Signatures is selected from the set $\left\{8 + i\right\}_{i=0}^{56}$. All G-Signatures hyperparameter searches are done in a non-exhaustive way.

        \begin{table}[h!]
            \begin{center}
                \begin{tabular}{l c c c c}
                \toprule
              {\makecell{\bf {Method}}} & {\makecell{\bf \#{}Layers}} & {\makecell{\bf Hidden Size}} & {\makecell{\bf Learning Rate}} & {\makecell{\bf \#{}Heads}} \\
                \midrule
                    GGCNs & $4$ & $70$ & $1.00 \times 10^{-3}$ & - \\
                    GTs & $10$ & $80$ & $1.00 \times 10^{-3}$ & $8$ \\
                    G.-Sigs. & $1$ & $73$ & $3.30 \times 10^{-3}$ & $2$ \\
                \bottomrule
                \end{tabular}
            \end{center}
            \caption{Final hyperparameter settings for compared methods on CoMA. For G-Signatures, the final signature size is set to $63$, and the final weight decay is set to $1.68 \times 10^{-5}$.}
        \end{table}

        \pagebreak
        
        \begin{table}[ht]
            \begin{center}
                \begin{tabular}{l c c c c c}
                \toprule
              {\makecell{\bf {Method}}} & {\makecell{\bf \#{}Layers}} & {\makecell{\bf Hidden Size}} & {\makecell{\bf Learning Rate}} & {\makecell{\bf \#{}Heads}} & {\makecell{\bf \#{}Modes}}\\
                \midrule
                    ResNets & $8$ & $128$ & $1.00 \times 10^{-2}$ & - & - \\
                    FNOs & $4$ & $128$ & $1.00 \times 10^{-2}$  & - & $12$ \\
                    G.-Sigs. & $1$ & $84$ & $1.05 \times 10^{-3}$ & $2$ & - \\
                \bottomrule
                \end{tabular}
            \end{center}
            \caption{Final hyperparameter settings for compared methods on the KPZ dataset with high noise. For G-Signatures, the final signature size is set to $27$, and the final weight decay is set to $1.31 \times 10^{-6}$.}
        \end{table}
                \begin{table}[ht]
            \begin{center}
                \begin{tabular}{l c c c c c}
                \toprule
              {\makecell{\bf {Method}}} & {\makecell{\bf \#{}Layers}} & {\makecell{\bf Hidden Size}} & {\makecell{\bf Learning Rate}} & {\makecell{\bf \#{}Heads}} & {\makecell{\bf \#{}Modes}}\\
                \midrule
                    ResNets & $8$ & $128$ & $1.00 \times 10^{-2}$ & - & - \\
                    FNOs & $4$ & $128$ & $1.00 \times 10^{-2}$ & - & $12$ \\
                    G.-Sigs. & $1$ & $89$ & $1.19 \times 10^{-3}$ & $3$ & - \\
                \bottomrule
                \end{tabular}
            \end{center}
            \caption{Final hyperparameter settings for compared methods on the KPZ dataset with low noise. For G-Signatures, the final signature size is set to $47$, and the final weight decay is set to $2.47 \times 10^{-4}$.}
        \end{table}
        \begin{table}[ht]
            \begin{center}
                \begin{tabular}{l c c c c c}
                \toprule
              {\makecell{\bf {Method}}} & {\makecell{\bf \#{}Layers}} & {\makecell{\bf Hidden Size}} & {\makecell{\bf Learning Rate}} & {\makecell{\bf \#{}Heads}} & {\makecell{\bf \#{}Modes}}\\
                \midrule
                    ResNets & $8$ & $128$ & $1.00 \times 10^{-2}$ & - & - \\
                    FNOs & $4$ & $128$ & $1.00 \times 10^{-2}$ & - & $12$ \\
                    G.-Sigs. & $1$ & $81$ & $1.01 \times 10^{-3}$ & $1$ & - \\
                \bottomrule
                \end{tabular}
            \end{center}
            \caption{Final hyperparameter settings for compared methods on the KPZ dataset with no noise. For G-Signatures, the final signature size is set to $61$, and the final weight decay is set to $8.46 \times 10^{-4}$.}
        \end{table}

        \begin{table}[h!]
            \begin{center}
                \begin{tabular}{l c c c c}
                \toprule
              {\makecell{\bf {Method}}} & {\makecell{\bf \#{}Layers}} & {\makecell{\bf Hidden Size}} & {\makecell{\bf Learning Rate}} & {\makecell{\bf \#{}Heads}}\\
                \midrule
                    GGCNs & $8$ & $70$ & $1.00 \times 10^{-3}$ & - \\
                    GTs & $6$ & $32$ & $1.00 \times 10^{-3}$ & $8$ \\
                    G.-Sigs. & $1$ & $86$ & $5.87 \times 10^{-3}$ & $3$ \\
                \bottomrule
                \end{tabular}
            \end{center}
            \caption{Final hyperparameter settings for compared methods on the ETA dataset with 500 nodes and a sparsity of $0.0$. For G-Signatures, the final signature size is set to $35$, and the final weight decay is set to $2.55 \times 10^{-4}$.}
        \end{table}
        
        \begin{table}[h!]
            \begin{center}
                \begin{tabular}{l c c c c}
                \toprule
              {\makecell{\bf {Method}}} & {\makecell{\bf \#{}Layers}} & {\makecell{\bf Hidden Size}} & {\makecell{\bf Learning Rate}} & {\makecell{\bf \#{}Heads}}\\
                \midrule
                    GGCNs & $10$ & $70$ & $1.00 \times 10^{-3}$ & - \\
                    GTs & $8$ & $32$ & $1.00 \times 10^{-3}$ & $8$ \\
                    G.-Sigs. $1$ & $95$ & $3.32 \times 10^{-3}$ & $3$ \\
                \bottomrule
                \end{tabular}
            \end{center}
            \caption{Final hyperparameter settings for compared methods on the ETA dataset with 500 nodes and a sparsity of $0.5$. For G-Signatures, the final signature size is set to $43$, and the final weight decay is set to $1.82 \times 10^{-6}$.}
        \end{table}

        \pagebreak

        \begin{table}[h!]
            \begin{center}
                \begin{tabular}{l c c c c}
                \toprule
              {\makecell{\bf {Method}}} & {\makecell{\bf \#{}Layers}} & {\makecell{\bf Hidden Size}} & {\makecell{\bf Learning Rate}} & {\makecell{\bf \#{}Heads}}\\
                \midrule
                    GGCNs & $10$ & $70$ & $1.00 \times 10^{-3}$ & - \\
                    GTs & $8$ & $32$ & $1.00 \times 10^{-3}$ & $8$ \\
                    G.-Sigs. & $1$ & $32$ & $2.84 \times 10^{-3}$ & $2$ \\
                \bottomrule
                \end{tabular}
            \end{center}
            \caption{Final hyperparameter settings for compared methods on the ETA dataset with 500 nodes and a sparsity of $0.9$. For G-Signatures, the final signature size is set to $27$, and the final weight decay is set to $3.39 \times 10^{-5}$.}
        \end{table}

        \begin{table}[ht]
            \begin{center}
                \begin{tabular}{l c c c c}
                \toprule
              {\makecell{\bf {Method}}} & {\makecell{\bf \#{}Layers}} & {\makecell{\bf Hidden Size}} & {\makecell{\bf Learning Rate}} & {\makecell{\bf \#{}Heads}}\\
                \midrule
                    GGCNs & $8$ & $70$ & $1.00 \times 10^{-3}$ & - \\
                    GTs & $6$ & $32$ & $1.00 \times 10^{-3}$ & $8$ \\
                    G.-Sigs. & $1$ & $79$ & $2.91 \times 10^{-3}$ & $3$ \\
                \bottomrule
                \end{tabular}
            \end{center}
            \caption{Final hyperparameter settings for compared methods on the ETA dataset with 1000 nodes and a sparsity of $0.0$. For G-Signatures, the final signature size is set to $40$, and the final weight decay is set to $1.82 \times 10^{-6}$.}
        \end{table}

        \begin{table}[ht]
            \begin{center}
                \begin{tabular}{l c c c c}
                \toprule
              {\makecell{\bf {Method}}} & {\makecell{\bf \#{}Layers}} & {\makecell{\bf Hidden Size}} & {\makecell{\bf Learning Rate}} & {\makecell{\bf \#{}Heads}}\\
                \midrule
                    GGCNs & $10$ & $70$ & $1.00 \times 10^{-3}$ & - \\
                    GTs & $8$ & $32$ & $1.00 \times 10^{-3}$ & $8$ \\
                    G.-Sigs. & $1$ & $71$ & $3.51 \times 10^{-3}$ & $3$ \\
                \bottomrule
                \end{tabular}
            \end{center}
            \caption{Final hyperparameter settings for compared methods on the ETA dataset with 1000 nodes and a sparsity of $0.5$. For G-Signatures, the final signature size is set to $26$, and the final weight decay is set to $4.15 \times 10^{-5}$.}
        \end{table}

        \begin{table}[h!]
            \begin{center}
                \begin{tabular}{l c c c c}
                \toprule
              {\makecell{\bf {Method}}} & {\makecell{\bf \#{}Layers}} & {\makecell{\bf Hidden Size}} & {\makecell{\bf Learning Rate}} & {\makecell{\bf \#{}Heads}}\\
                \midrule
                    GGCNs & $10$ & $70$ & $1.00 \times 10^{-3}$ & - \\
                    GTs & $8$ & $32$ & $1.00 \times 10^{-3}$ & $8$ \\
                    G.-Sigs. & $1$ & $50$ & $4.13 \times 10^{-3}$ & $2$ \\
                \bottomrule
                \end{tabular}
            \end{center}
            \caption{Final hyperparameter settings for compared methods on the ETA dataset with 1000 nodes and a sparsity of $0.9$. For G-Signatures, the final signature size is set to $40$, and the final weight decay is set to $1.75 \times 10^{-4}$.}
        \end{table}

        \begin{table}[h!]
            \begin{center}
                \begin{tabular}{l c c c c}
                \toprule
              {\makecell{\bf {Method}}} & {\makecell{\bf \#{}Layers}} & {\makecell{\bf Hidden Size}} & {\makecell{\bf Learning Rate}} & {\makecell{\bf \#{}Heads}}\\
                \midrule
                    GGCNs & $8$ & $100$ & $1.00 \times 10^{-3}$ & - \\
                    GTs GTs & $6$ & $80$ & $1.00 \times 10^{-3}$ & $8$ \\
                    G.-Sigs. & $1$ & $75$ & $4.81 \times 10^{-3}$ & $3$ \\
                \bottomrule
                \end{tabular}
            \end{center}
            \caption{Final hyperparameter settings for compared methods on the ETA dataset with 2000 nodes and a sparsity of $0.0$. For G-Signatures, the final signature size is set to $64$, and the final weight decay is set to $2.29 \times 10^{-6}$.}
        \end{table}

        \pagebreak

        \begin{table}[ht]
            \begin{center}
                \begin{tabular}{l c c c c}
                \toprule
              {\makecell{\bf {Method}}} & {\makecell{\bf \#{}Layers}} & {\makecell{\bf Hidden Size}} & {\makecell{\bf Learning Rate}} & {\makecell{\bf \#{}Heads}}\\
                \midrule
                    GGCNs & $10$ & $100$ & $1.00 \times 10^{-3}$ & - \\
                    GTs GTs & $8$ & $80$ & $1.00 \times 10^{-3}$ & $8$ \\
                    G.-Sigs. & $1$ & $59$ & $5.06 \times 10^{-3}$ & $2$ \\
                \bottomrule
                \end{tabular}
            \end{center}
            \caption{Final hyperparameter settings for compared methods on the ETA dataset with 2000 nodes and a sparsity of $0.5$. For G-Signatures, the final signature size is set to $53$, and the final weight decay is set to $1.95 \times 10^{-6}$.}
        \end{table}

        \begin{table}[ht]
            \begin{center}
                \begin{tabular}{l c c c c}
                \toprule
              {\makecell{\bf {Method}}} & {\makecell{\bf \#{}Layers}} & {\makecell{\bf Hidden Size}} & {\makecell{\bf Learning Rate}} & {\makecell{\bf \#{}Heads}}\\
                \midrule
                    GGCNs & $10$ & $100$ & $1.00 \times 10^{-3}$ & - \\
                    GTs GTs & $8$ & $80$ & $1.00 \times 10^{-3}$ & $8$ \\
                    G.-Sigs. & $1$ & $77$ & $2.70 \times 10^{-3}$ & $2$ \\
                \bottomrule
                \end{tabular}
            \end{center}
            \caption{Final hyperparameter settings for compared methods on the ETA dataset with 2000 nodes and a sparsity of $0.9$. For G-Signatures, the final signature size is set to $36$, and the final weight decay is set to $2.05 \times 10^{-6}$.}
        \end{table}

    \section{Acknowledgements}
        
        The authors thank Markus Holzleitner for discussions on rough paths theory, and for pointing us towards important aspects of theorems concerning randomized signatures.
        
        The ELLIS Unit Linz, the LIT AI Lab, the Institute for Machine Learning, are supported by the Federal State Upper Austria. IARAI is supported by Here Technologies. We thank the projects AI- MOTION (LIT-2018-6-YOU-212), AI-SNN (LIT-2018-6-YOU-214), DeepFlood (LIT-2019-8-YOU- 213), Medical Cognitive Computing Center (MC3), INCONTROL-RL (FFG-881064), PRIMAL (FFG-873979), S3AI (FFG-872172), DL for GranularFlow (FFG-871302), AIRI FG 9-N (FWF- 36284, FWF-36235), ELISE (H2020-ICT-2019-3 ID: 951847). We thank Audi.JKU Deep Learning Center, TGW LOGISTICS GROUP GMBH, Silicon Austria Labs (SAL), FILL Gesellschaft mbH, Anyline GmbH, Google, ZF Friedrichshafen AG, Robert Bosch GmbH, UCB Biopharma SRL, Merck Healthcare KGaA, Verbund AG, Software Competence Center Hagenberg GmbH, TÜV Austria, Frauscher Sensonic and the NVIDIA Corporation.

\end{document}